\documentclass[10pt,letterpaper]{article}
\usepackage[textwidth=5.5in,textheight=9in,centering]{geometry}
\usepackage[T1]{fontenc}
\usepackage{newunicodechar}
\newunicodechar{’}{\textquoteright}
\usepackage{times}
\usepackage[round,authoryear]{natbib}
\usepackage{microtype}
\usepackage{titlesec}
\titleformat{\section}{\large\bfseries}{\thesection}{1em}{}
\titleformat{\subsection}{\normalsize\bfseries}{\thesubsection}{1em}{}
\titleformat{\subsubsection}{\normalsize\itshape}{\thesubsubsection}{1em}{}
\usepackage{graphicx}
\usepackage{xcolor}

\usepackage{amsmath}
\usepackage{amssymb}
\usepackage{bm}

\definecolor{tickgreen}{RGB}{35,165,90}
\definecolor{crossred}{RGB}{160,55,55}
\newcommand{\cmark}{\textcolor{tickgreen}{\ensuremath{\bm{\checkmark}}}}
\newcommand{\xmark}{\textcolor{crossred}{\ensuremath{\bm{\times}}}}

\usepackage{booktabs}
\usepackage{multirow}
\usepackage{array}
\newcommand{\metriccell}[2]{%
  \shortstack[c]{\ensuremath{#1}\\[-0.2ex]%
  \scalebox{0.78}{\ensuremath{(\pm #2)}}}%
}

\usepackage{placeins}

\usepackage{url}
\usepackage[hidelinks]{hyperref}
\hypersetup{
  pdftitle={PPCAR-Net: Projection-Refined Parametric 3D Coronary Artery Reconstruction from Sparse X-ray Angiographic Views},
  pdfauthor={Yu Ren, Hwee Kuan Lee, Tat-Jen Cham, Jonathan Yap, Khung Keong Yeo}
}

\graphicspath{{figures/}}

\title{PPCAR-Net: Projection-Refined Parametric 3D Coronary Artery Reconstruction from Sparse X-ray Angiographic Views}

\author{%
  \normalsize Yu Ren\textsuperscript{1,2}, Hwee Kuan Lee\textsuperscript{1}, Tat-Jen Cham\textsuperscript{2},\\
  \normalsize Jonathan Yap\textsuperscript{3}, Khung Keong Yeo\textsuperscript{3}\\[0.6em]
  \small \textsuperscript{1}Bioinformatics Institute, Agency for Science, Technology and Research (A*STAR)\\
  \small \textsuperscript{2}College of Computing and Data Science, Nanyang Technological University\\
  \small \textsuperscript{3}National Heart Centre Singapore
}
\date{}

\newcommand{\codeurl}{https://github.com/G2304138H/PPCAR-Net}
\newcommand{\codelink}{%
  \if\relax\detokenize\expandafter{\codeurl}\relax
    \textit{[GitHub URL to be added]}%
  \else
    \expandafter\url\expandafter{\codeurl}%
  \fi
}

\begin{document}

\maketitle

\begin{abstract}
Sparse-view 3D coronary reconstruction commonly relies on cross-view correspondence and triangulation, which are vulnerable to vessel overlap and foreshortening, or on volumetric prediction followed by vascular-graph extraction, which does not directly provide centrelines and radii. We introduce PPCAR-Net, a projection-refined parametric coronary artery reconstruction network that directly predicts a branch-structured centreline-and-radius representation without explicit point matching, triangulation, or an intermediate volume. Given a variable number of segmented views, a coarse predictor combines frozen VGGT features with learned branch queries to estimate branch presence, B-spline centreline trajectories, and dense radius profiles. Projection-guided geometry and radius refiners then sample local evidence from the input views and apply residual corrections learned with 3D supervision. We evaluate representation fidelity and sparse-view reconstruction quantitatively and qualitatively. On simulated angiographic masks generated from CT-derived coronary anatomy, PPCAR-Net produces better connected artery reconstructions and achieves strong centreline accuracy, particularly for RCA, while maintaining competitive volumetric overlap. Coarse-to-fine inference takes $121$ ms, enabling real-time reconstruction. Code and models are publicly available at: \codelink.
\end{abstract}


\section{Introduction}
\label{sec:introduction}

X-ray coronary angiography is fundamental for diagnosing coronary artery disease and guiding intervention. Each angiogram, however, represents a two-dimensional projection of inherently three-dimensional coronary anatomy. Foreshortening and vessel overlap can obscure spatial relationships between branches. In clinical practice, simultaneous biplane acquisitions are uncommon, and complementary monoplane views are generally acquired sequentially. Consequently, cardiologists must mentally integrate multiple views to infer three-dimensional vessel geometry~\citep{klein1998quantitative}. Conventional 3D quantitative coronary angiography partially automates this process by reconstructing anatomy from two calibrated, phase-matched projections through cross-view vessel correspondence and triangulation~\citep{tu2010biplane}. However, sparse projections, overlapping branches, and foreshortening can make these correspondences ambiguous. We therefore investigate whether population-level patterns of coronary anatomy learned across training cases can support a plausible branch-structured 3D hypothesis when the projection evidence is ambiguous.

Existing learning-based methods commonly represent coronary arteries using neural fields, Gaussian primitives, or voxel volumes, rather than directly producing ordered coronary centrelines and lumen radii. Extracting vascular graphs from volumetric outputs can introduce errors that compromise downstream analysis and interpretation. We instead adopt the dense centreline-and-radius representation~\citep{iyer2023multistage} as the model’s native output. We parameterise each dense branch centreline as a continuous cubic B-spline to ensure connectivity within each branch, while retaining dense radius values for localised radius variation. We refer to this branch-structured B-spline centreline-and-radius representation as \emph{vessel code}. It is directly interpretable and editable, without requiring vascular-graph extraction from a volumetric prediction.

\begin{table*}[t]
\centering
\caption{Comparison of learning-based coronary artery reconstruction methods. Per-case runtime is the complete time required to reconstruct one 3D artery from two input views, including vascular-graph extraction when applicable.}
\label{tab:method-capability-comparison}
\scriptsize
\setlength{\tabcolsep}{3pt}
\begin{tabular*}{\textwidth}{@{\extracolsep{\fill}}>{\raggedright\arraybackslash}m{0.22\textwidth}>{\centering\arraybackslash}m{0.08\textwidth}>{\raggedright\arraybackslash}m{0.24\textwidth}>{\centering\arraybackslash}m{0.055\textwidth}>{\centering\arraybackslash}m{0.10\textwidth}>{\centering\arraybackslash}m{0.055\textwidth}>{\centering\arraybackslash}m{0.065\textwidth}@{}}
\toprule
\multirow{2}{*}{\raisebox{-1.2ex}{\textbf{Method}}}
& \multicolumn{1}{>{\raggedright\arraybackslash}m{0.08\textwidth}}{\multirow{2}{*}{\raisebox{-1.2ex}{\shortstack[l]{\textbf{Number of}\\\textbf{input views}}}}}
& \multirow{2}{*}{\raisebox{-1.2ex}{\shortstack[l]{\textbf{Output}\\\textbf{representation}}}}
& \multicolumn{3}{c}{\textbf{Output}}
& \multicolumn{1}{l}{\multirow{2}{*}{\raisebox{-1.2ex}{\shortstack[l]{\textbf{Per-case}\\\textbf{runtime} $\downarrow$}}}} \\
\cmidrule(lr){4-6}
& & & \shortstack{\textbf{3D}\\\textbf{volume}} & \shortstack{\textbf{Connected}\\\textbf{centreline}} & \textbf{Radius} & \\
\midrule
\citet{iyer2023multistage} & 2 or 3 & Dense centreline points and radii & \cmark & \cmark & \cmark & -- \\
NeRF-CA~\citep{maas2024nerfca} & $\geq 4$ & Novel 2D views & \xmark & \xmark & \xmark & -- \\
NerT-CA~\citep{maas2026nertca} & $\geq 3$ & Novel 2D views & \xmark & \xmark & \xmark & -- \\
SDF-CAR~\citep{reda2026sdfcar} & 2 & 3D implicit SDF field & \cmark & \xmark & \xmark & 45 min \\
3DGR-CAR~\citep{fu20243dgrcar} & $\geq 2$ & 3D Gaussian primitives & \cmark & \xmark & \xmark & 63.7 s \\
DeepCA~\citep{wang2025deepca} & 2 & 3D voxel & \cmark & \xmark & \xmark & 2.69 s \\
AutoCAR~\citep{zhu2025autocar} & 2 & 3D voxel & \cmark & \xmark & \xmark & 3.76 s \\
PPCAR-Net (ours) & $\geq 1$ & B-spline centreline curves and dense radii & \cmark & \cmark & \cmark & 121 ms \\
\bottomrule
\end{tabular*}
\end{table*}

We propose PPCAR-Net, a projection-refined parametric coronary artery reconstruction network that predicts vessel code through a coarse-to-fine process. Given a variable number of 2D coronary masks and their acquisition directions, a coarse predictor combines frozen VGGT features with learned branch queries to estimate branch existence, B-spline control points, and dense radius profiles. It infers a plausible branch-structured 3D hypothesis: the branch-existence predictions determine the number of active branches, the B-spline control points describe their trajectories, and the dense-radius predictions provide approximate radius profiles. Projection-guided geometry and radius refiners then sample local 2D evidence and apply learned residual corrections to the centreline geometry and radii. This feed-forward design empirically improves reconstruction accuracy without per-case optimisation, enabling sub-second inference. We train and evaluate using simulated angiographic masks generated from the ImageCAS CT angiography dataset~\citep{zeng2023imagecas}. Qualitative examples on diseased ASOCA anatomy~\citep{gharleghi2023annotated} and preprocessed real X-ray angiography explore anatomical and image-domain transfer, respectively. Table~\ref{tab:method-capability-comparison} compares the capabilities of our approach with representative learning-based methods.

The contributions of this work are: 
\begin{itemize}
    \item We demonstrate that integrating a coronary anatomical prior learned across the training cohort with projection-guided corrections based on case-specific evidence reconstructs accurate, anatomically structured 3D coronary arteries from sparse views, without explicit point correspondence or per-case optimisation. These results suggest that learned anatomical priors provide useful constraints under ambiguity from sparse X-ray angiographic views.
    
    \item PPCAR-Net, supported by branch-subset augmentation during training, achieves the best RCA structural and centreline reconstruction metrics among evaluated methods, with competitive volumetric accuracy and robust branch prediction. This is accomplished with sub-second inference, demonstrating both high accuracy and real-time capability.
    
    \item We enhance the explicit branch-structured centreline-and-radius representation by incorporating a smooth cubic B-spline parameterisation of dense polyline centrelines. This adaptation leads to more coherent and accurate vessel geometry reconstructions compared to dense point predictions, while maintaining interpretability and intra-branch continuity.
    
\end{itemize}

\section{Related Work}
\label{sec:related-work}

Conventional 3D quantitative coronary angiography reconstructs coronary geometry from calibrated X-ray projections by segmenting target vessels, establishing cross-view centreline correspondence, and triangulating matched points~\citep{klein1998quantitative,tu2010biplane}. Related parametric approaches have used deformable B-splines for 2D vessel-boundary and centreline analysis~\citep{klein1997deformable}, 3D B-spline curves for motion tracking in biplane angiograms~\citep{shechter2003motion}, and NURBS surfaces for structured coronary reconstruction and meshing~\citep{vukicevic2018nurbs}. These pipelines may require user interaction and remain sensitive to vessel overlap and foreshortening, motivating learning-based reconstruction from segmented angiograms.

Recently, learning-based methods have been used for automating the mapping of angiographic images to 3D vessels. AutoCAR backprojects image features into a sparse 3D volume and predicts vascular occupancy and centreness; post-processing converts this prediction into a vascular graph, while iterative deformation relates graphs reconstructed at contracted and relaxed cardiac states~\citep{zhu2025autocar}. DeepCA follows a related volumetric strategy: it backprojects two segmented, non-simultaneous projections into a 3D input volume and uses a conditional generative network to reconstruct a voxelised coronary tree while implicitly compensating for inter-view motion~\citep{wang2025deepca}. In contrast, \citet{iyer2023multistage} directly predict an ordered centreline-and-radius representation from segmented, uncalibrated angiograms and their model is trained and evaluated primarily on synthetically generated RCA trees with a fixed number of branches. More broadly, VesselDiffusion generates vascular point sets using a diffusion model~\citep{guo2025vesseldiffusion}. Because paired 2D projections and corresponding 3D artery labels are rarely available in clinical X-ray angiography, supervised methods commonly generate paired training data from coronary CT angiography and evaluate image-domain transfer on real X-ray angiograms.

Neural-field and Gaussian representations have also been applied to coronary reconstruction. NeRF-CA and NerT-CA model dynamic coronary appearance within an X-ray attenuation field and synthesise new composite X-ray-like views~\citep{maas2024nerfca,maas2026nertca}. 3DGR-CAR represents coronary occupancy with Gaussian primitives and combines a learned initialisation with projection-based parameter optimisation~\citep{fu20243dgrcar}, whereas SDF-CAR fits a patient-specific neural implicit field to two segmented projections~\citep{reda2026sdfcar}. Although these methods recover or imply 3D structure, their native neural-field, Gaussian, and occupancy representations do not directly estimate explicit coronary artery geometries for downstream tasks such as blood-flow simulation; extracting arterial structure through post-processing may introduce additional errors and does not leverage learned structural priors.

\section{Methods}
\label{sec:methods}

\subsection{3D Coronary Artery Reconstruction Task}
\label{subsec:task}

Our task is to reconstruct an explicit 3D centreline-and-radius representation of a coronary artery from a variable number of calibrated 2D coronary mask projections. For each coronary case, the input is the variable-sized set
\begin{equation}
\mathcal{X}
=
\left\{
(\bm{I}_k,\theta_k,\phi_k)
\right\}_{k=1}^{K},
\end{equation}
where $K$ is the number of available views, $\bm{I}_k$ is the coronary artery mask in view $k$, and $(\theta_k,\phi_k)$ specifies its acquisition direction. Each view follows a cone-beam projection model
$\Pi_k:\mathbb{R}^{3}\rightarrow\mathbb{R}^{2}$, whose pose is determined by $(\theta_k,\phi_k)$ and whose fixed intrinsic parameters map 3D coordinates to detector pixels. We use the same camera model to generate the input projections and to guide the refinement described in later subsections. The target centreline-and-radius representation is introduced next.

\begin{figure*}[t]
\centering
\includegraphics[width=0.96\textwidth]{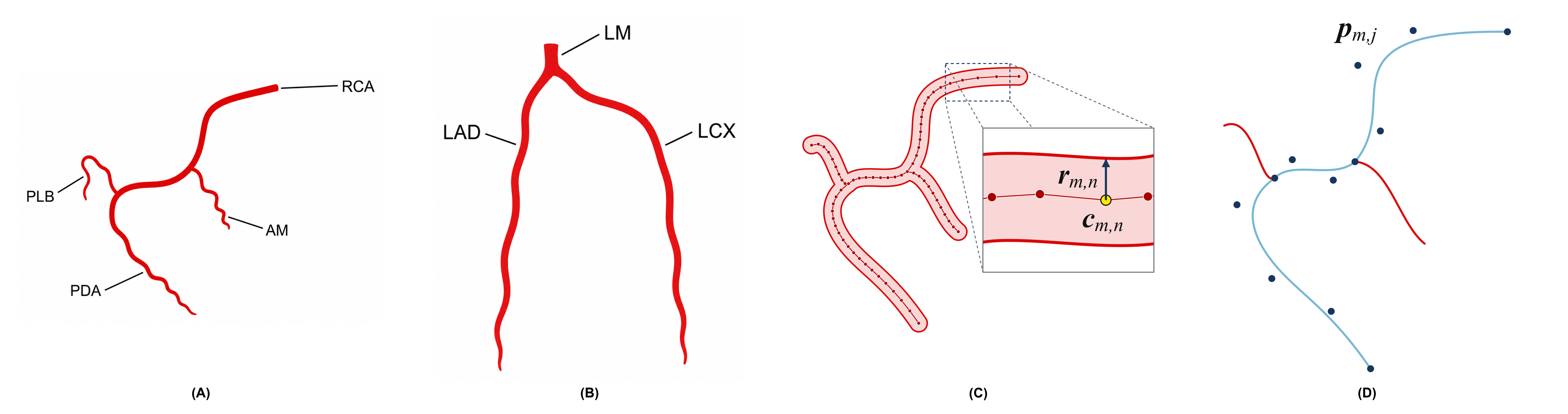}
\caption{Coronary trees and vessel code representation. (A) Schematic RCA tree showing the right coronary artery (RCA), posterolateral branch (PLB), posterior descending artery (PDA), and acute marginal branch (AM); (B) Schematic LCA tree showing the left main (LM), left anterior descending (LAD), and left circumflex (LCX) arteries; (C) Dense centreline $\bm{c}_{m,n}$ and radius $r_{m,n}$ representation of a branch; (D) B-spline representation of the centreline of branch $m$ with dark-blue points denoting B-spline control points $\bm{p}_{m,j}$ and the light-blue centreline that they define.}
\label{fig:parametric-vessel-representation}
\end{figure*}

\subsection{Parametric Vessel Representation and Dataset Construction}
\label{subsec:parametric-representation}

Here, we formalise \emph{vessel code}, the explicit branch-structured centreline-and-radius representation predicted by PPCAR-Net, and describe how its training targets are constructed. A complete coronary tree is represented using at most $M$ branches, indexed by $m\in\{1,\ldots,M\}$, with $e_m\in\{0,1\}$ indicating whether branch $m$ is present. For each existing branch, we adopt the dense centreline-and-radius representation of \citet{iyer2023multistage}. Branch $m$ with $e_m=1$ is described by $N$ ordered centreline-and-radius samples running from its proximal root or attachment point on its parent branch to its distal endpoint, as illustrated in Figure~\ref{fig:parametric-vessel-representation}(C):
\begin{equation}
\bm{V}_m
=
\{(\bm{c}_{m,n},r_{m,n})\}_{n=1}^{N},
\qquad
\bm{c}_{m,n}\in\mathbb{R}^{3},
\quad
r_{m,n}\in\mathbb{R}_{>0}.
\end{equation}
Here, $\bm{c}_{m,n}$ denotes the $n$-th centreline point of branch $m$ in 3D space, and $r_{m,n}$ denotes the vessel radius associated with that centreline point. A circular lumen profile centred at $\bm{c}_{m,n}$ with radius $r_{m,n}$ provides the corresponding tubular surface approximation.

To better reflect the continuous structure of an artery while reducing the output dimensionality, we parameterise the raw centreline coordinates by a cubic B-spline with $K_{\mathrm{cp}}$ control points, and the reconstruction network predicts these control points directly. For branch $m$, the control-point matrix $\bm{P}_m\in\mathbb{R}^{K_{\mathrm{cp}}\times3}$ and the decoded dense centreline points are
\begin{equation}
\bm{P}_m
=
(\bm{p}_{m,1},\ldots,\bm{p}_{m,K_{\mathrm{cp}}})^{\mathsf T},
\qquad
\bm{c}_{m,n}
=
\sum_{j=1}^{K_{\mathrm{cp}}}
B_{j,3}(t_n)\bm{p}_{m,j},
\quad
t_n=\frac{n-1}{N-1}.
\end{equation}
Here, $\bm{p}_{m,j}\in\mathbb{R}^{3}$ is the $j$-th control point, $t_n$ is the normalised arc-length position of centreline point $n$, and $B_{j,3}(\cdot)$ is the $j$-th cubic B-spline basis function defined by the Cox--de Boor recursion. As shown in Figure~\ref{fig:parametric-vessel-representation}(D), the control points determine the decoded centreline but do not need to lie on it. Because a B-spline is a continuous curve over its parameter domain, the decoded centreline of every branch is connected by construction.

Radius is deliberately not compressed by the centreline spline. The dense vector
$
\bm{r}_m
=
(r_{m,1},\ldots,r_{m,N})^{\mathsf T}
\in
\mathbb{R}_{>0}^{N}
$
retains one radius at every decoded centreline point. The complete vessel code is therefore $\{(e_m,\bm{P}_m,\bm{r}_m)\}_{m=1}^{M}$. This parameterisation provides a low-dimensional smooth prior for centreline geometry without imposing the same smoothness bottleneck on localised radius changes.

\paragraph{Dataset and target construction.}
\label{subsec:dataset}
Supervised training uses coronary annotations from the ImageCAS CT angiography dataset~\citep{zeng2023imagecas} as 3D labels. Each annotation is separated into RCA and LCA components (Figure~\ref{fig:parametric-vessel-representation}(A--B)). Each component mask is skeletonised to extract ordered branch centrelines, and the radius at each centreline point is estimated using the Euclidean distance transform, obtaining an ordered branch-wise vessel code. Branches follow deterministic appearance-based positions in the target sequence, allowing optional side branches without assigning every query a fixed anatomical name. Each retained branch contains $N=200$ proximal-to-distal centreline-and-radius samples, represented by $K_{\mathrm{cp}}=20$ B-spline control points and dense radii in millimetres. The target control points are fitted offline by endpoint-constrained least squares, and $K_{\mathrm{cp}}=20$ yields an overall mean centreline fitting error of $0.164$ mm and a centreline-based overlap score of $99.92\%$ between the fitted and original vessels (Appendix Table~\ref{tab:bspline-fitting-fidelity}). This process yields 750 RCA and 750 LCA components, with each cohort divided into 600 training, 75 validation, and 75 test cases. For every ImageCAS case, the cone-beam model $\Pi_k$ generates seven $256\times256$ synthetic coronary-mask projections that mimic segmented X-ray angiograms at clinically motivated acquisition directions, from which the requested number of views is selected during training and evaluation. Separately, the 20 diseased CT angiography cases from ASOCA~\citep{gharleghi2023annotated} are used only for qualitative analysis; their annotations are not converted into vessel code targets or included in training or quantitative evaluation.

\subsection{Flexible-View Coarse Parametric Predictor}
\label{subsec:coarse-predictor}

\begin{figure*}[t]
\centering
\includegraphics[width=\textwidth]{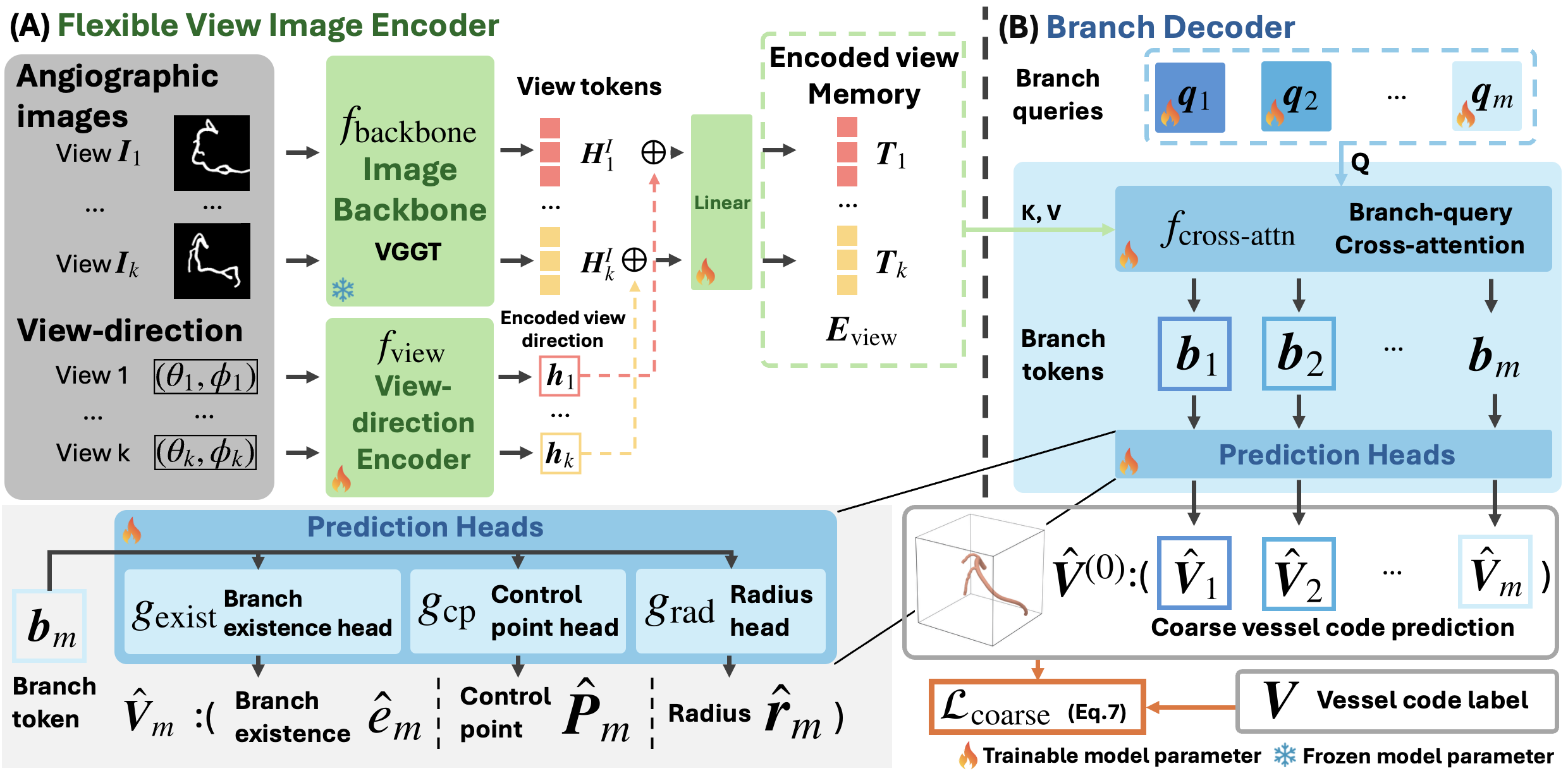}
\caption{Flexible-view coarse parametric predictor in PPCAR-Net: (A) Frozen VGGT features are concatenated ($\oplus$) with encoded acquisition directions to form the multiview memory $\bm{E}_{\mathrm{view}}$. (B) Trainable branch queries cross-attend to this memory, and the existence, control-point, and radius heads produce the coarse vessel code $\hat{\bm{V}}^{(0)}$. Fire and snowflake symbols denote trainable and frozen modules, respectively.}
\label{fig:coarse-model-overview}
\end{figure*}

The first stage of PPCAR-Net is a flexible-view coarse parametric predictor, summarised in Figure~\ref{fig:coarse-model-overview}. The coarse predictor combines evidence from the available views with learned patterns of coronary anatomy to infer a plausible structured vessel code. The existence predictions determine the active number of branches, while the control-point and radius heads estimate their global trajectories and approximate radius profiles.

As illustrated in Figure~\ref{fig:coarse-model-overview}(A), the frozen VGGT image backbone $f_{\mathrm{backbone}}$ extracts image tokens $\bm{H}^{I}_k$ from each input image $\bm{I}_k$, while the view-direction encoder $f_{\mathrm{view}}$ encodes its acquisition direction $(\theta_k,\phi_k)$ as $\bm{h}_k$. To obtain the view-direction-aware tokens $\bm{T}_k$, the same vector $\bm{h}_k$ is repeated for every image token, concatenated with $\bm{H}^{I}_k$ along the feature dimension, and passed through a learned projection back to the model dimension $d$. The tokens from all available views form the multiview memory
$
\bm{E}_{\mathrm{view}}
=
[\bm{T}_1;\ldots;\bm{T}_K].
$
The number of views $K$ and their acquisition directions can vary between cases.

Figure~\ref{fig:coarse-model-overview}(B) shows how $M$ trainable branch queries $\bm{q}_m$ retrieve branch-specific evidence from this memory. The query matrix $\bm{Q}_{\mathrm{query}}=[\bm{q}_1;\ldots;\bm{q}_M]$ cross-attends to $\bm{E}_{\mathrm{view}}$ through $f_{\mathrm{cross\text{-}attn}}$, with $\bm{Q}_{\mathrm{query}}$ serving as the queries and $\bm{E}_{\mathrm{view}}$ as the keys and values. This produces branch tokens $[\bm{b}_1;\ldots;\bm{b}_M]$, where $\bm{b}_m\in\mathbb{R}^{d}$. The branch queries correspond to the ordered branch entries: $\bm{q}_m$ retrieves image evidence for the $m$-th branch entry, while the existence head determines whether that branch is present in the current case. The resulting branch token $\bm{b}_m$ is mapped by the existence, control-point, and radius
heads to the coarse vessel code
$
\hat{\bm{V}}^{(0)}
=
\left\{
(\hat e_m,\hat{\bm{P}}^{(0)}_m,\hat{\bm{r}}^{(0)}_m)
\right\}_{m=1}^{M},
$
where $\hat e_m$, $\hat{\bm{P}}^{(0)}_m$, and $\hat{\bm{r}}^{(0)}_m$ denote
the predicted branch existence, B-spline control points, and dense radii,
respectively. This stage learns the mapping from input appearance to 3D artery structure through paired training data. It therefore produces a plausible prior-guided 3D hypothesis but does not enforce direct projection-space consistency. The following refinement stage uses input view evidence to correct this hypothesis.

\subsection{Projection-Guided Refinement}
\label{subsec:projection-guided-refinement}

\begin{figure*}[t]
\centering
\includegraphics[width=\textwidth]{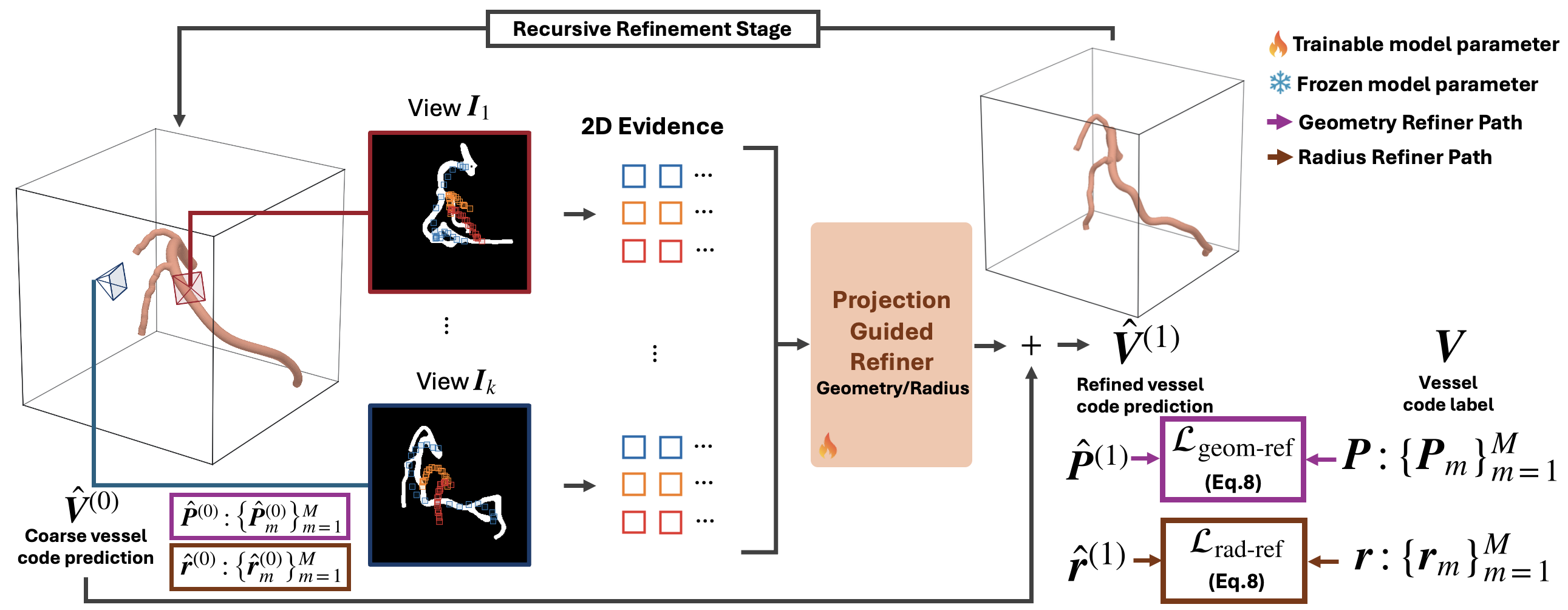}
\caption{Recursive projection-guided refinement in PPCAR-Net. The B-spline geometry refiner uses local multiview evidence to predict residual updates to the coarse control points. With the refined centreline fixed, the dense-radius refiner then updates the pointwise radii. The two refiners are trained sequentially and separately, using $\mathcal{L}_\mathrm{geom\text{-}ref}$ and $\mathcal{L}_\mathrm{rad\text{-}ref}$, respectively.}
\label{fig:projection-guided-refinement-overview}
\end{figure*}

The refinement component of PPCAR-Net, summarised in Figure~\ref{fig:projection-guided-refinement-overview}, progressively updates the coarse vessel code $\hat{\bm{V}}^{(0)}$ using projection-guided geometry and radius corrections. At each refinement stage, the current centreline and radii are projected into every available input view using the same cone-beam operator $\Pi_k$ employed to generate the input projections. Local 2D evidence is sampled around these projected positions and aggregated across views through cross-attention. A residual head then predicts a correction to the current 3D parameter. 

For the geometry refiner, at refinement stage $s$, the query associated with control point $\hat{\bm{p}}^{(s)}_{m,j}$ cross-attends to the multiview evidence sampled around its corresponding projected centreline position. The resulting geometry residual $\Delta\hat{\bm{p}}^{(s)}_{m,j}$ updates the control point as
$
\hat{\bm{p}}^{(s+1)}_{m,j}
=
\hat{\bm{p}}^{(s)}_{m,j}
+
\Delta\hat{\bm{p}}^{(s)}_{m,j}
$. The updated B-spline is decoded and reprojected before the next stage samples new evidence. We use $S_g=4$ shared-parameter geometry-refinement stages while holding the branch-existence predictions and coarse radii fixed. After geometry refinement, the dense-radius refiner applies the same residual principle at every decoded centreline point. A radius query cross-attends to the local multiview evidence and predicts
$\Delta\hat r^{(s)}_{m,n}$, giving
$
\hat r^{(s+1)}_{m,n}
=
\hat r^{(s)}_{m,n}
+
\Delta\hat r^{(s)}_{m,n}
$. The refined centreline geometry is held fixed throughout this module. We use $S_r=3$ shared-parameter radius-refinement stages, producing the final vessel code $\hat{\bm{V}}_{\mathrm{refined}}$.

\subsection{Staged Training Objectives}
\label{subsec:training-signals}

PPCAR-Net is trained sequentially: first the coarse predictor, then the geometry refiner with the coarse predictor frozen, and finally the radius refiner with the preceding modules frozen. Because target branches have a deterministic ordering, prediction $m$ is compared directly with ground-truth branch $m$. At prediction stage $s$, the geometry loss supervises both the B-spline control points and the centreline decoded from them:
\begin{align}
\mathcal{L}_{\mathrm{geometry}}^{(s)}
={}&
\frac{\lambda_{\mathrm{cp}}}{N_{\mathrm{cp}}}
\sum_{m,j}e_{m}
\left\lVert
\hat{\bm{p}}^{(s)}_{m,j}-\bm{p}_{m,j}
\right\rVert_2^2
+
\frac{\lambda_{\mathrm{cl}}}{N_{\mathrm{cl}}}
\sum_{m,n}e_{m}
\left\lVert
\hat{\bm{c}}^{(s)}_{m,n}-\bm{c}_{m,n}
\right\rVert_2^2,
\label{eq:geometry-loss-main}
\end{align}
where $N_{\mathrm{cp}}$ and $N_{\mathrm{cl}}$ denote the total numbers of control points and decoded centreline points across existing branches, and $\lambda_{\mathrm{cp}}$ and $\lambda_{\mathrm{cl}}$ weight the corresponding supervision terms. While the B-spline parameterisation guarantees intra-branch connectivity, the attachment loss encourages connectivity between side branches and their parent branch. Let $\kappa_{m}\in\{1,\ldots,N\}$ denote the ground-truth index of this junction on the parent branch centreline. We regularise the predicted junction by
\begin{equation}
\mathcal{L}_{\mathrm{attach}}^{(s)}
=
\frac{1}{\sum_me_m}
\sum_me_{m}
\left\lVert
\hat{\bm{c}}^{(s)}_{m,1}
-
\hat{\bm{c}}^{(s)}_{p,\kappa_{m}}
\right\rVert_2^2,
\label{eq:attachment-loss-main}
\end{equation}
where $p$ denotes the index of the parent branch. This loss directly encourages the predicted first point of each existing side branch to coincide with its corresponding predicted parent branch attachment point. The radius loss is the pointwise mean absolute error at corresponding dense centreline points:
\begin{equation}
\mathcal{L}_{\mathrm{radius}}^{(s)}
=
\frac{1}{N_{\mathrm{cl}}}
\sum_{m,n}e_{m}
\left|
\hat r^{(s)}_{m,n}-r_{m,n}
\right|.
\end{equation}
Across these losses, $e_{m}$ excludes absent branches. For the coarse prediction, $s=0$, and binary cross-entropy $\mathcal{L}_{\mathrm{exist}}$ supervises the predicted branch-existence labels:
\begin{equation}
\mathcal{L}_{\mathrm{coarse}}
=
\mathcal{L}_{\mathrm{geometry}}^{(0)}
+
\lambda_{\mathrm{attach}}
\mathcal{L}_{\mathrm{attach}}^{(0)}
+
\lambda_{\mathrm{radius}}
\mathcal{L}_{\mathrm{radius}}^{(0)}
+
\lambda_{\mathrm{exist}}
\mathcal{L}_{\mathrm{exist}},
\end{equation}
Absent branch entries therefore contribute to existence supervision but not to geometry, attachment, or radius supervision. For refinement, the losses used are
\begin{equation}
    \mathcal{L}_{\mathrm{geom\text{-}ref}}
    =
    \sum_{s=1}^{S_g}\alpha_s(\mathcal{L}_{\mathrm{geometry}}^{(s)}+\lambda_{\mathrm{attach}}\mathcal{L}_{\mathrm{attach}}^{(s)}),
    \qquad
    \mathcal{L}_{\mathrm{rad\text{-}ref}}=\sum_{s=1}^{S_r}\beta_s\mathcal{L}_{\mathrm{radius}}^{(s)},
\end{equation}
where $\mathcal{L}_{\mathrm{geom\text{-}ref}}$ collects geometry and attachment losses over the recursive geometry stages, while $\mathcal{L}_{\mathrm{rad\text{-}ref}}$ collects radius losses over the radius stages. Here $\alpha_s$ and $\beta_s$ are loss weights. These objectives are applied separately: $\mathcal{L}_{\mathrm{geom\text{-}ref}}$ trains only the geometry refiner, whereas $\mathcal{L}_{\mathrm{rad\text{-}ref}}$ trains only the radius refiner. The same existence and valid-point masks are applied throughout refinement. 


\paragraph{Within-case branch-subset augmentation training.}
To model optional side branches, each training tree is paired with variants formed by successively removing side branches in reverse canonical order until only the required main path remains. Updated projections, existence labels, and geometry targets from all variants train the coarse predictor and geometry refiner, varying branch presence while retaining the same underlying anatomy.

\subsection{Evaluation Metrics}
\label{subsec:evaluation-metrics}

We evaluate reconstruction accuracy using Dice, centreline Dice (clDice), Chamfer distance (CD), and connected-component count (CC count). Dice measures the volumetric overlap between the predicted and ground-truth 3D arteries and is reported as a percentage, making it sensitive to both vessel location and lumen extent. For volumetric evaluation, the prediction and ground-truth artery are represented on the same isotropic grid with a voxel spacing of $0.5\,\mathrm{mm}$ along each axis. clDice measures topology-aware agreement between the skeletons of the voxelised prediction and ground-truth artery. CD measures the geometric discrepancy in millimetres between the predicted and ground-truth centreline point sets. Each evaluated RCA or LCA tree forms one connected vascular structure because every retained side branch attaches to its corresponding main branch; consequently, its ground-truth volume contains exactly one connected component. To evaluate prediction fragmentation, each method's native output is converted to the same voxel grid and its connected foreground components are counted. The ideal CC count is therefore one, while larger values indicate disconnected predicted artery segments. Dice and clDice provide complementary volumetric and centreline-topology assessments, CD assesses centreline localisation, and CC count directly evaluates whether the prediction preserves the expected connected artery structure. Higher Dice and clDice values, a lower CD, and a lower CC count indicate better reconstruction. Every metric is first calculated per case and then summarised over the test cases with mean and standard error.

Additional data-construction and architectural details are provided in Appendix Sections~\ref{suppsec:dataset-construction} and~\ref{suppsec:model-training-protocols}, respectively.

\section{Results}
\label{sec:results}

\subsection{Comparison model selection}

We quantitatively compare PPCAR-Net with 3DGR-CAR~\citep{fu20243dgrcar}, AutoCAR~\citep{zhu2025autocar}, and DeepCA~\citep{wang2025deepca}; the latter two are evaluated only in their fixed two-view setting. NeRF-CA~\citep{maas2024nerfca} and NerT-CA~\citep{maas2026nertca} generate novel X-ray-like views rather than 3D arteries; the method of \citet{iyer2023multistage} operates on uncalibrated input views and assumes a fixed number of artery branches; therefore these methods are not evaluated. SDF-CAR reconstructs a 3D artery without cross-case pre-training, but its per-patient optimisation requires approximately 45 minutes per case~\citep{reda2026sdfcar}, and thus is excluded from quantitative comparison.

\subsection{RCA and LCA Reconstruction}
\label{subsec:main-results}
Table~\ref{tab:reconstruction-results} evaluates reconstruction from one, two, and four input views on the 75-case RCA test split described in Section~\ref{subsec:parametric-representation}, using Dice, clDice, Chamfer distance, and connected-component count. For methods without a native centreline graph, we extract the centreline using the post-processing approach of \citet{zhu2025autocar}. All methods use the same view subsets and evaluation protocol. 

Compared with the other methods in Table~\ref{tab:reconstruction-results}, our refined model outperforms 3DGR-CAR in the one-view setting. Under the conventional two-view setting, AutoCAR achieves the highest Dice, whereas our method achieves the highest clDice and lowest Chamfer distance, indicating stronger centreline-topology agreement and localisation despite the small difference in volumetric overlap. With four views, our refined model substantially outperforms 3DGR-CAR across Dice, clDice, and Chamfer distance. Dice alone does not fully capture vessel connectivity because disconnected foreground regions may still overlap with the target artery. As each branch centreline is connected by construction and attachment supervision encourages side branches to join the main branch, our refined predictions achieve the lowest CC count at every evaluated view setting and remain close to the ideal value of one, while the comparison methods produce more fragmented outputs.

\begin{table*}[t]
\centering
\caption{RCA reconstruction from one, two, and four input views. CD denotes Chamfer distance, and CC count denotes connected-component count. Values are mean $\pm$ standard error. The best result is shown in \textbf{bold} and the second-best result is \underline{underlined}. -- denotes a view-count setting that is not reported in the corresponding paper.}
\label{tab:reconstruction-results}
\scriptsize
\setlength{\tabcolsep}{2pt}
\begin{tabular*}{\textwidth}{@{\extracolsep{\fill}}
>{\raggedright\arraybackslash}m{0.15\textwidth}
*{12}{>{\centering\arraybackslash}m{0.057\textwidth}}@{}}
\toprule
\multirow{2}{*}{\textbf{Method}}
& \multicolumn{4}{c}{\textbf{One view}}
& \multicolumn{4}{c}{\textbf{Two views}}
& \multicolumn{4}{c}{\textbf{Four views}} \\
\cmidrule(lr){2-5}\cmidrule(lr){6-9}\cmidrule(lr){10-13}
& \shortstack{\textbf{Dice}\\\textbf{(\%)} $\uparrow$} & \shortstack{\textbf{clDice}\\\textbf{(\%)} $\uparrow$} & \shortstack{\textbf{CD}\\\textbf{(mm)} $\downarrow$} & \shortstack{\textbf{CC}\\\textbf{count} $\downarrow$}
& \shortstack{\textbf{Dice}\\\textbf{(\%)} $\uparrow$} & \shortstack{\textbf{clDice}\\\textbf{(\%)} $\uparrow$} & \shortstack{\textbf{CD}\\\textbf{(mm)} $\downarrow$} & \shortstack{\textbf{CC}\\\textbf{count} $\downarrow$}
& \shortstack{\textbf{Dice}\\\textbf{(\%)} $\uparrow$} & \shortstack{\textbf{clDice}\\\textbf{(\%)} $\uparrow$} & \shortstack{\textbf{CD}\\\textbf{(mm)} $\downarrow$} & \shortstack{\textbf{CC}\\\textbf{count} $\downarrow$} \\
\midrule
\shortstack[l]{3DGR-CAR\\\citep{fu20243dgrcar}}
& \metriccell{\underline{22.18}}{0.65} & \metriccell{\underline{22.35}}{0.73} & \metriccell{\underline{10.11}}{0.44} & \metriccell{72.77}{12.08}
& \metriccell{37.32}{0.91} & \metriccell{39.47}{1.09} & \metriccell{7.72}{0.48} & \metriccell{159.62}{16.84}
& \metriccell{\underline{44.33}}{1.12} & \metriccell{\underline{46.54}}{1.31} & \metriccell{\underline{7.37}}{0.49} & \metriccell{216.25}{17.95} \\
\shortstack[l]{AutoCAR\\\citep{zhu2025autocar}}
& -- & -- & -- & --
& \metriccell{\mathbf{68.76}}{0.70} & \metriccell{\underline{79.13}}{1.37} & \metriccell{\underline{3.24}}{0.27} & \metriccell{3.35}{0.31}
& -- & -- & -- & -- \\
\shortstack[l]{DeepCA\\\citep{wang2025deepca}}
& -- & -- & -- & --
& \metriccell{63.49}{1.56} & \metriccell{78.00}{1.73} & \metriccell{3.78}{0.51} & \metriccell{4.41}{0.60}
& -- & -- & -- & -- \\
\midrule
\shortstack[l]{PPCAR-Net\\(coarse)}
& \metriccell{15.73}{1.46} & \metriccell{16.43}{1.84} & \metriccell{11.86}{0.59} & \metriccell{\underline{1.324}}{0.072}
& \metriccell{19.47}{1.12} & \metriccell{20.40}{1.56} & \metriccell{9.70}{0.41} & \metriccell{\underline{1.324}}{0.064}
& \metriccell{21.45}{1.26} & \metriccell{22.88}{1.70} & \metriccell{9.16}{0.36} & \metriccell{\underline{1.284}}{0.059} \\
\shortstack[l]{PPCAR-Net\\(refined)}
& \metriccell{\mathbf{27.37}}{1.86} & \metriccell{\mathbf{32.38}}{2.56} & \metriccell{\mathbf{9.39}}{0.55} & \metriccell{\mathbf{1.081}}{0.032}
& \metriccell{\underline{66.92}}{1.50} & \metriccell{\mathbf{80.62}}{1.92} & \metriccell{\mathbf{3.16}}{0.27} & \metriccell{\mathbf{1.054}}{0.026}
& \metriccell{\mathbf{73.05}}{1.25} & \metriccell{\mathbf{88.54}}{1.45} & \metriccell{\mathbf{2.12}}{0.19} & \metriccell{\mathbf{1.014}}{0.014} \\
\bottomrule
\end{tabular*}
\end{table*}

\begin{table*}[t]
\centering
\caption{LCA reconstruction from one, two, and four input views. CD denotes Chamfer distance, and CC count denotes connected-component count. Values are mean $\pm$ standard error. The best result is shown in \textbf{bold} and the second-best result is \underline{underlined}. -- denotes a view-count setting that is not reported in the corresponding paper.}
\label{tab:supp-lca-reconstruction-results}
\scriptsize
\setlength{\tabcolsep}{2pt}
\begin{tabular*}{\textwidth}{@{\extracolsep{\fill}}
>{\raggedright\arraybackslash}m{0.15\textwidth}
*{12}{>{\centering\arraybackslash}m{0.057\textwidth}}@{}}
\toprule
\multirow{2}{*}{\textbf{Method}}
& \multicolumn{4}{c}{\textbf{One view}}
& \multicolumn{4}{c}{\textbf{Two views}}
& \multicolumn{4}{c}{\textbf{Four views}} \\
\cmidrule(lr){2-5}\cmidrule(lr){6-9}\cmidrule(lr){10-13}
& \shortstack{\textbf{Dice}\\\textbf{(\%)} $\uparrow$} & \shortstack{\textbf{clDice}\\\textbf{(\%)} $\uparrow$} & \shortstack{\textbf{CD}\\\textbf{(mm)} $\downarrow$} & \shortstack{\textbf{CC}\\\textbf{count} $\downarrow$}
& \shortstack{\textbf{Dice}\\\textbf{(\%)} $\uparrow$} & \shortstack{\textbf{clDice}\\\textbf{(\%)} $\uparrow$} & \shortstack{\textbf{CD}\\\textbf{(mm)} $\downarrow$} & \shortstack{\textbf{CC}\\\textbf{count} $\downarrow$}
& \shortstack{\textbf{Dice}\\\textbf{(\%)} $\uparrow$} & \shortstack{\textbf{clDice}\\\textbf{(\%)} $\uparrow$} & \shortstack{\textbf{CD}\\\textbf{(mm)} $\downarrow$} & \shortstack{\textbf{CC}\\\textbf{count} $\downarrow$} \\
\midrule
\shortstack[l]{3DGR-CAR\\\citep{fu20243dgrcar}}
& \metriccell{\mathbf{26.51}}{0.51} & \metriccell{\mathbf{26.48}}{0.56} & \metriccell{\mathbf{8.25}}{0.17} & \metriccell{46.43}{3.54}
& \metriccell{35.05}{0.48} & \metriccell{37.16}{0.54} & \metriccell{\mathbf{5.87}}{0.11} & \metriccell{76.10}{8.06}
& \metriccell{\underline{43.64}}{0.66} & \metriccell{\underline{47.81}}{0.75} & \metriccell{\mathbf{4.94}}{0.20} & \metriccell{100.88}{11.59} \\
\shortstack[l]{AutoCAR\\\citep{zhu2025autocar}}
& -- & -- & -- & --
& \metriccell{46.89}{1.29} & \metriccell{\underline{63.00}}{1.59} & \metriccell{6.13}{0.38} & \metriccell{11.36}{0.78}
& -- & -- & -- & -- \\
\shortstack[l]{DeepCA\\\citep{wang2025deepca}}
& -- & -- & -- & --
& \metriccell{\mathbf{56.56}}{0.81} & \metriccell{\mathbf{71.20}}{1.11} & \metriccell{6.66}{0.24} & \metriccell{9.97}{0.32}
& -- & -- & -- & -- \\
\midrule
\shortstack[l]{PPCAR-Net\\(coarse)}
& \metriccell{11.15}{0.76} & \metriccell{10.10}{0.81} & \metriccell{14.08}{0.51} & \metriccell{\underline{1.829}}{0.118}
& \metriccell{14.92}{0.77} & \metriccell{13.94}{0.90} & \metriccell{12.37}{0.51} & \metriccell{\underline{1.605}}{0.097}
& \metriccell{14.59}{0.73} & \metriccell{14.05}{0.93} & \metriccell{12.46}{0.49} & \metriccell{\underline{1.618}}{0.095} \\
\shortstack[l]{PPCAR-Net\\(refined)}
& \metriccell{\underline{19.73}}{1.00} & \metriccell{\underline{21.36}}{1.19} & \metriccell{\underline{11.31}}{0.45} & \metriccell{\mathbf{1.197}}{0.050}
& \metriccell{\underline{49.45}}{1.32} & \metriccell{61.30}{1.93} & \metriccell{\underline{5.98}}{0.47} & \metriccell{\mathbf{1.053}}{0.026}
& \metriccell{\mathbf{53.45}}{1.28} & \metriccell{\mathbf{67.43}}{1.93} & \metriccell{\underline{5.30}}{0.42} & \metriccell{\mathbf{1.039}}{0.022} \\
\bottomrule
\end{tabular*}
\end{table*}

Table~\ref{tab:supp-lca-reconstruction-results} reports reconstruction from the same view counts on the 75-case LCA test split, using the same metrics and evaluation protocol. The LCA task is more challenging than RCA because its representation supports up to 13 branches, increasing anatomical complexity, projected branch overlap, and ambiguity under sparse views. In the conventional two-view setting, DeepCA achieves the highest Dice and clDice, while our method has lower Chamfer distance than AutoCAR and DeepCA and substantially fewer connected components ($1.053$ versus $11.36$ and $9.97$, respectively). AutoCAR and DeepCA backproject the observed views into volumetric representations, which can favour agreement with the input silhouettes and voxel-overlap metrics. In contrast, the connected-component results and the qualitative examples in Figure~\ref{fig:3d-reconstruction-comparison} show that
our explicit representation more consistently preserves a coherent artery-tree structure.  Under one-view input, 3DGR-CAR achieves higher overlap scores but produces $46.43$ connected components, compared with $1.197$ for our method. Its predicted foreground volume is also substantially larger than the ground truth, indicating that part of its overlap arises from broad, fragmented coverage. Although a single projection does not determine depth, our model still produces a nearly connected artery-shaped 3D hypothesis rather than a fragmented volume. With four views, our method outperforms 3DGR-CAR in Dice, clDice, and connectivity, although its Chamfer distance remains higher.

\subsection{Ablation Studies}
\label{subsec:refinement-results}

Our ablation studies examine the contributions of projection-guided refinement, B-spline parameterisation, and branch-subset augmentation in PPCAR-Net. Starting from the same coarse prediction, direct per-scene optimisation improves reconstruction accuracy but underperforms learned refinement across all four reconstruction metrics for both RCA and LCA (Table~\ref{tab:refinement-ablation}). Direct optimisation fits only the observed 2D masks, which may admit geometrically implausible solutions under sparse-view ambiguity. In contrast, the learned refiners use 3D supervision across training cases to predict residual corrections from projection evidence. The combined geometry and radius refiner achieves $66.92\%$ RCA Dice, compared with $41.02\%$ for direct optimisation, while requiring only $61$ ms rather than $11$ s after coarse prediction.

B-spline parameterisation and branch-subset augmentation further improve reconstruction. Replacing direct dense-centreline prediction with B-spline control-point prediction increases two-view RCA refined Dice from $22.18\%$ to $66.92\%$ (Appendix Table~\ref{tab:centreline-representation-ablation}). Branch-subset augmentation increases geometry-refined Dice from $46.57\%$ to $65.89\%$, branch-existence accuracy from $92.95\%$ to $94.63\%$, and F1 from $88.37\%$ to $92.48\%$, supporting more reliable identification of optional side branches (Appendix Table~\ref{tab:progressive-branch-augmentation}).

\begin{table*}[t]
\centering
\caption{Refinement ablation in PPCAR-Net from a common two-view coarse initialisation. CD denotes Chamfer distance, and CC count denotes connected-component count. Values are mean $\pm$ standard error; post-coarse time excludes the shared coarse prediction. The best result is shown in \textbf{bold} and the second-best result is \underline{underlined}.}
\label{tab:refinement-ablation}
\scriptsize
\setlength{\tabcolsep}{2pt}
\begin{tabular*}{\textwidth}{@{\extracolsep{\fill}}
>{\raggedright\arraybackslash}m{0.18\textwidth}
*{8}{>{\centering\arraybackslash}m{0.072\textwidth}}
>{\centering\arraybackslash}m{0.10\textwidth}@{}}
\toprule
\multirow{2}{*}{\textbf{Reconstruction stage}}
& \multicolumn{4}{c}{\textbf{RCA}}
& \multicolumn{4}{c}{\textbf{LCA}}
& \multirow{2}{*}{\shortstack{\textbf{Post-coarse}\\\textbf{time} $\downarrow$}} \\
\cmidrule(lr){2-5}\cmidrule(lr){6-9}
& \shortstack{\textbf{Dice}\\\textbf{(\%)} $\uparrow$}
& \shortstack{\textbf{clDice}\\\textbf{(\%)} $\uparrow$}
& \shortstack{\textbf{CD}\\\textbf{(mm)} $\downarrow$}
& \shortstack{\textbf{CC}\\\textbf{count} $\downarrow$}
& \shortstack{\textbf{Dice}\\\textbf{(\%)} $\uparrow$}
& \shortstack{\textbf{clDice}\\\textbf{(\%)} $\uparrow$}
& \shortstack{\textbf{CD}\\\textbf{(mm)} $\downarrow$}
& \shortstack{\textbf{CC}\\\textbf{count} $\downarrow$}
& \\
\midrule
Coarse prediction
& \metriccell{19.47}{1.12} & \metriccell{20.40}{1.56} & \metriccell{9.70}{0.41} & \metriccell{1.324}{0.064}
& \metriccell{14.92}{0.77} & \metriccell{13.94}{0.90} & \metriccell{12.37}{0.51} & \metriccell{1.605}{0.097} & -- \\
Direct per-scene optimisation
& \metriccell{41.02}{1.62} & \metriccell{50.02}{2.24} & \metriccell{6.65}{0.39} & \metriccell{1.243}{0.057}
& \metriccell{28.75}{1.12} & \metriccell{33.16}{1.51} & \metriccell{9.29}{0.50} & \metriccell{1.421}{0.073} & 11 s \\
Learned geometry refiner
& \metriccell{\underline{65.89}}{1.44} & \metriccell{\mathbf{80.84}}{1.90} & \metriccell{\mathbf{3.16}}{0.27} & \metriccell{\underline{1.068}}{0.029}
& \metriccell{\underline{48.43}}{1.29} & \metriccell{\mathbf{61.38}}{1.93} & \metriccell{\mathbf{5.98}}{0.47} & \metriccell{\mathbf{1.053}}{0.026} & \textbf{14 ms} \\
Learned geometry and radius refiners
& \metriccell{\mathbf{66.92}}{1.50} & \metriccell{\underline{80.62}}{1.92} & \metriccell{\underline{3.16}}{0.27} & \metriccell{\mathbf{1.054}}{0.026}
& \metriccell{\mathbf{49.45}}{1.32} & \metriccell{\underline{61.30}}{1.93} & \metriccell{\underline{5.98}}{0.47} & \metriccell{\underline{1.053}}{0.026} & \underline{61 ms} \\
\bottomrule
\end{tabular*}
\end{table*}

\subsection{Qualitative Reconstruction and Transfer to Real X-ray Angiography}

\begin{figure*}[t]
\centering
\IfFileExists{figures/reconstruction_3d_comparison.png}{%
    \includegraphics[width=\textwidth]{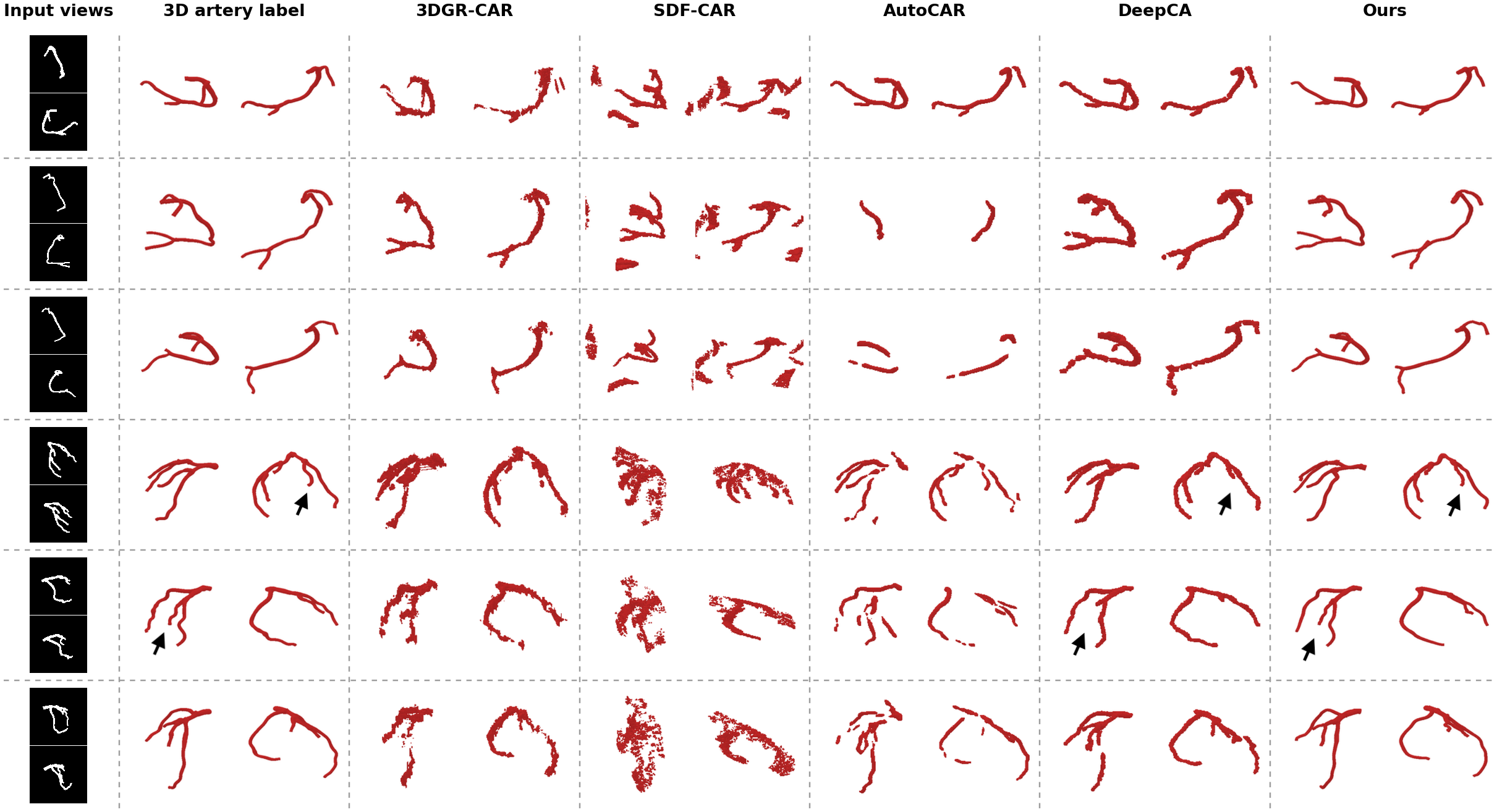}%
}{%
    \fbox{\parbox[c][38mm][c]{0.94\textwidth}{\centering
    Placeholder for normal RCA reconstructions and selected RCA stenosis cases.}}%
}
\caption{Qualitative two-view reconstruction. Columns show the input masks, 3D label, and the method outputs; rows show three RCA cases and three LCA cases, each rendered from two viewpoints. Arrows mark branches poorly visible in the input projections.}
\label{fig:3d-reconstruction-comparison}
\end{figure*}

\begin{figure*}[t]
\centering
\IfFileExists{figures/figure5_sparse_view_reconstruction.png}{%
    \includegraphics[width=\textwidth]{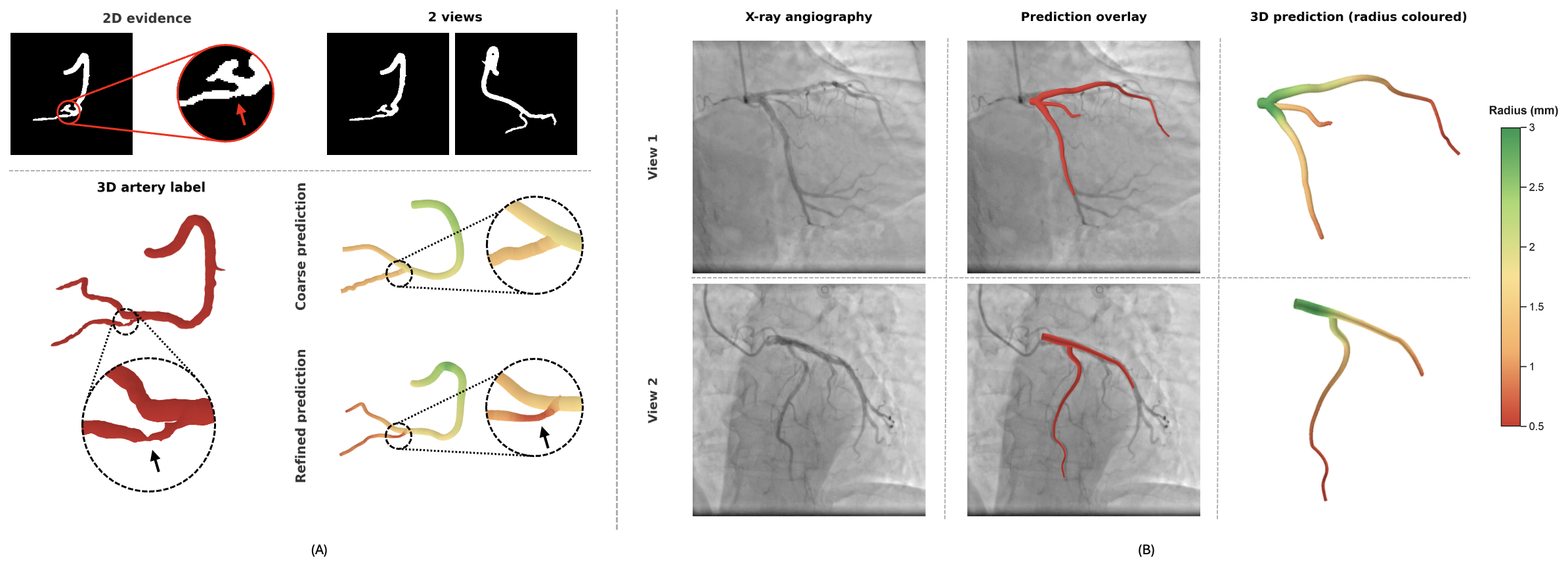}%
}{%
    \fbox{\parbox[c][42mm][c]{0.75\textwidth}{\centering
    Placeholder: qualitative one-, two-, and four-view reconstruction.}}%
}
\caption{Qualitative reconstruction with PPCAR-Net on diseased CT anatomy and real X-ray angiography. (A) A diseased ASOCA~\citep{gharleghi2023annotated} case with a visually identified region of local artery narrowing. The 3D label, simulated masks, and radius-coloured coarse and refined predictions are shown. (B) A real two-view X-ray angiography case from the AutoCAR dataset~\citep{zhu2025autocar}. The middle column overlays the prediction on each angiographic view, and the final column shows the radius-coloured 3D prediction.}
\label{fig:sparse-view-reconstruction}
\end{figure*}

Figure~\ref{fig:3d-reconstruction-comparison} qualitatively illustrates these connectivity differences by showing gaps in voxel and implicit predictions. Subsequent graph extraction may fail to fill the gaps in the predicted artery. In contrast, our continuous branch representation preserves within-branch connectivity and exposes branch attachments and radii. The arrows in Figure~\ref{fig:3d-reconstruction-comparison} mark branches that are weakly visible in the inputs but recovered by our model, suggesting that learned anatomical patterns complement sparse projection evidence. Figure~\ref{fig:sparse-view-reconstruction}(A) shows that the coarse prediction captures the global artery shape, while projection-guided refinement uses 2D evidence to recover the corresponding radius reduction. This indicates sensitivity to visible narrowing, rather than clinical stenosis detection. Figure~\ref{fig:sparse-view-reconstruction}(B) demonstrates application to preprocessed real angiography, although the absence of paired 3D labels prevents accuracy assessment. Additional qualitative video results are available on our project page, accessible through the GitHub repository.

Implementation details and comparison-method adaptations are described in Appendix Sections~\ref{suppsec:experiment-details} and~\ref{suppsec:comparison-method-implementation}, respectively. An additional ablation of the frozen image backbone and an inference-time breakdown for the comparison methods are reported in Appendix Tables~\ref{tab:backbone-two-view-comparison} and~\ref{tab:supp-inference-time-results}.

\section{Discussion and Conclusion}
\label{sec:discussion}

We introduce PPCAR-Net, a coarse-to-fine network that combines learned coronary anatomical patterns with projection-guided corrections to predict an explicit vessel code. Its explicit vessel-code representation guarantees a connected centreline within each branch, while attachment supervision produces predictions close to a single connected artery tree, a desirable property for clinical interpretation and downstream geometric analysis. The feed-forward pipeline enables sub-second reconstruction without per-case optimisation.

Figure~\ref{fig:sparse-view-reconstruction}(B) provides an initial qualitative application to real X-ray coronary angiography, but it does not constitute systematic clinical validation. Sequentially acquired clinical monoplane views can retain cardiac deformation even after phase matching. Future work should evaluate larger clinical cohorts and address inter-view misalignment caused by motion and gantry displacement~\citep{tu2010biplane} through motion-aware reconstruction and geometry calibration. A central open question is how to evaluate 3D reconstruction accuracy without a 3D artery label; paired acquisitions of CT and X-ray coronary angiography or expert assessment may therefore be required.

The learned coronary anatomy patterns can support a plausible reconstruction when projection evidence is incomplete, but they cannot reliably identify a specific stenosis that is concealed by overlap or foreshortening in every available view. Moreover, without clinical stenosis annotations, our current examples assess radius reconstruction rather than diagnostic stenosis detection. Further studies using expert lesion annotations and physiological measurements such as fractional flow reserve are therefore required to determine whether reconstructed radius changes correspond to clinically significant stenoses.

Finally, the present model requires known 3D artery labels for supervised training. Developing unsupervised or self-supervised objectives from real angiographic views could reduce this dependence; the current model may also provide initial pseudo-3D labels for datasets that contain only 2D X-ray coronary angiography images.

\clearpage
\FloatBarrier
\bibliography{references}
\bibliographystyle{plainnat}

\clearpage
\appendix
\section{Appendix}
\label{sec:appendix}

\subsection{Dataset and Target-Construction Details}
\label{suppsec:dataset-construction}

The ImageCAS coronary annotations~\citep{zeng2023imagecas} are provided as NIfTI volumes. We use the $x$-, $y$-, and $z$-axis spacing metadata to express all centreline coordinates, radii, and distances in millimetres. Starting from the 1,000 source cases, each labelled artery volume is automatically separated into RCA and LCA components and every separated component is inspected manually. A component is excluded if its RCA/LCA identity cannot be determined reliably, if the two artery systems cannot be separated, or if annotation defects prevent construction of a valid target. The resulting RCA and LCA collections are component cohorts rather than disjoint patient cohorts: retained RCA and LCA components may originate from the same ImageCAS case, so their counts do not represent completely different patients. These exclusions may bias the retained cohorts towards cases with clearer artery annotations and should be considered when interpreting generalisation to the complete clinical population.

Retained components are skeletonised in physical coordinates and converted into an initial centreline graph using local node connections and a minimum spanning tree, following the broad graph-construction procedure used by AutoCAR~\citep{zhu2025autocar}. Because automatic conversion can fail for tortuous vessels, rapidly changing radii, and annotation defects, manual supervision removes extra branches, repairs erroneous connections, and verifies branch identity and ordering. Difficult geometry is retained whenever a reliable target can be produced after correction. For each side branch, the first centreline point is set to its attachment point on the corresponding main branch, explicitly encoding branch connectivity in the target vessel code. Radius is obtained from the distance transform of the CT artery mask, with each cross-section approximated by a circular lumen profile. Every branch is resampled from its proximal root or attachment to its distal endpoint to produce $N=200$ ordered centreline-and-radius samples and is fitted with $K_{\mathrm{cp}}=20$ B-spline control points. This quality-control process produces 750 RCA and 750 LCA cases, split at patient level into 600 training, 75 validation, and 75 test cases.

For the qualitative narrowing example in Figure~\ref{fig:sparse-view-reconstruction}(A), we additionally use one diseased case from the ASOCA dataset~\citep{gharleghi2023annotated}. ASOCA provides 40 expert-annotated CT coronary angiography volumes, comprising 20 healthy participants and 20 participants with confirmed coronary artery disease. This ASOCA case is used only for qualitative evaluation and is not included in model training or the quantitative test cohorts. 

For RCA targets, the RCA trunk is the first branch and retained side branches are sorted by the normalised arc-length position of their attachment point while traversing the RCA trunk from proximal to distal. For LCA targets, the LM--LAD path is first, the LCX path is second, and the remaining side branches are sorted first along LAD and then along LCX, again by proximal-to-distal attachment position. Every retained ImageCAS LCA case contains a distinct LM artery. The ordering gives each optional side branch a deterministic target entry without assigning it a fixed anatomical name. For RCA, we retained a maximum of 7 branches; for LCA, we retained a maximum of 13 branches. For each case of RCA or LCA, unused side-branch entries are zero-padded and assigned an existence label of zero.

Seven clinically motivated projection anchors are defined separately for RCA and LCA, as listed in Table~\ref{tab:supp-view-direction-anchors}. Here, RAO and LAO denote the primary right and left anterior oblique rotations, while CRA and CAU denote the secondary cranial and caudal angulations; AP denotes zero primary rotation. These values are anchor directions rather than the only admissible acquisition angles, so nearby angular settings belong to the same clinical view family. For both RCA and LCA, the primary and secondary angles are independently varied within $\pm10^{\circ}$ of each artery-specific anchor.

\paragraph{Patient coordinates and projection centre.}
Vessel coordinates use the source-volume left--anterior--superior (LAS) patient frame: the positive $x$, $y$, and $z$ axes point towards the patient's left, anterior, and superior directions, respectively. No additional standing--supine rotation is applied. For each case, the projection isocentre is placed at the centroid of the reference artery surface. The reference is the main RCA surface for RCA and the combined LM--LAD and LCX surfaces for LCA. If this centroid is $\bm{o}$ in millimetres, the camera receives the centred point $\bar{\bm{x}}=10^{-3}(\bm{x}-\bm{o})$ in metres, so the isocentre is the world origin $\bm{O}=(0,0,0)^\top$.

\paragraph{Clinical angles and camera rotation.}
We use a positive clinical primary angle for LAO and a negative angle for RAO; a positive secondary angle denotes CRA and a negative angle denotes CAU. AP has zero primary angle. The preprocessing converts these clinical angles $(a_k,b_k)$ to the stored projector angles $(\theta_k,\phi_k)$. The conversion is $\theta_k=\operatorname{wrap}(90^{\circ}-a_k)$ and $\phi_k=90^{\circ}-b_k$, where $\operatorname{wrap}$ maps an angle to $[-180^{\circ},180^{\circ})$. 

\begin{table*}[t]
\centering
\caption{Clinical acquisition-direction anchors used to generate the seven candidate projections for each artery. Each entry reports the primary RAO/LAO rotation followed by the secondary CRA/CAU angulation.}
\label{tab:supp-view-direction-anchors}
\small
\setlength{\tabcolsep}{9pt}
\begin{tabular}{cll}
\toprule
\textbf{View} & \textbf{RCA anchor} & \textbf{LCA anchor} \\
\midrule
1 & LAO $40^{\circ}$, CRA $10^{\circ}$ & RAO $25^{\circ}$, CAU $35^{\circ}$ \\
2 & RAO $75^{\circ}$, CRA $10^{\circ}$ & LAO $5^{\circ}$, CAU $30^{\circ}$ \\
3 & AP, CRA $25^{\circ}$                & AP, CAU $10^{\circ}$ \\
4 & RAO $20^{\circ}$, CRA $20^{\circ}$ & RAO $40^{\circ}$, CRA $25^{\circ}$ \\
5 & LAO $20^{\circ}$                    & LAO $20^{\circ}$, CRA $30^{\circ}$ \\
6 & RAO $75^{\circ}$, CAU $20^{\circ}$ & LAO $40^{\circ}$, CRA $20^{\circ}$ \\
7 & LAO $60^{\circ}$                    & LAO $5^{\circ}$, CRA $40^{\circ}$ \\
\bottomrule
\end{tabular}
\end{table*}


\paragraph{Source, detector, and projection.}
The cone-beam projector uses a $256\times256$-pixel flat detector, a source-to-detector distance of $0.90\,\mathrm{m}$, and a source-to-isocentre distance of $0.75\,\mathrm{m}$. The detector pixel spacing is $0.55\,\mathrm{mm/pixel}$ for RCA and $0.65\,\mathrm{mm/pixel}$ for LCA.

\subsection{Model Details, Training, and Evaluation Protocols}
\label{suppsec:model-training-protocols}

This section provides additional architectural, target-fitting, and training details for PPCAR-Net, together with the test-time optimisation baseline.

\subsubsection{B-spline Representation and Target Fitting}
\label{suppsec:bspline-representation}

For each artery branch $m$, the ordered dense centreline samples $\{\bm{c}_{m,n}\}_{n=1}^{N}$ are parameterised by a cubic B-spline with $K_{\mathrm{cp}}=20$ controls. The fixed basis matrix $\bm{A}\in\mathbb{R}^{N\times K_{\mathrm{cp}}}$ is evaluated at $t_n=(n-1)/(N-1)$ and has entries $A_{n,j}=B_{j,3}(t_n)$. The basis uses the open-uniform clamped knot vector $\bm{u}=(0,0,0,0,1/17,\ldots,16/17,1,1,1,1)$. The branch control matrix is fitted offline by endpoint-constrained least squares:
\begin{equation}
\begin{aligned}
\bm{P}^{\star}_m
={}&
\arg\min_{\bm{P}_m}
\left\|
\bm{C}_m-\bm{A}\bm{P}_m
\right\|_{F}^{2},\\
\text{subject to}\quad
\bm{A}_{1,:}\bm{P}_m={}&\bm{c}_{m,1},\\
\bm{A}_{N,:}\bm{P}_m={}&\bm{c}_{m,N},
\end{aligned}
\end{equation}
where $\bm{P}_m\in\mathbb{R}^{K_{\mathrm{cp}}\times3}$ is the control-point matrix, $\bm{C}_m\in\mathbb{R}^{N\times3}$ contains the dense target coordinates, and $\|\cdot\|_F$ is the Frobenius norm. The decoded approximation is $\widetilde{\bm{C}}_m=\bm{A}\bm{P}^{\star}_m$.

Before training the image-to-artery model, we test whether the compact spline can represent the accepted ground-truth vessels. For each candidate control count, target controls are fitted and decoded to $N=200$ centreline points. Fitting error is the mean paired Euclidean distance between decoded and original points at corresponding normalised-arc-length positions. Vessel clDice is computed using the unchanged dense pointwise radii, thereby isolating centreline-parameterisation fidelity. Table~\ref{tab:bspline-fitting-fidelity} quantifies the approximation--compactness trade-off. With $K_{\mathrm{cp}}=20$, the overall fitting error is $0.164\pm0.108$ mm and vessel clDice is $99.92\pm0.62\%$, using 60 centreline parameters per branch instead of 600 dense coordinates. Table~\ref{tab:centreline-representation-ablation} complements this offline analysis by evaluating whether the B-spline parameterisation also benefits learned two-view coarse prediction and projection-guided refinement relative to directly predicting and refining all $N=200$ dense centreline points.

\begin{table*}[t]
\centering
\caption{Offline B-spline representation fidelity. Entries are mean $\pm$ sample standard deviation. Centreline fitting error is reported in millimetres and vessel clDice as a percentage. The RCA and LCA statistics each use 750 cases. The overall statistic pools 1,500 source-case means, with each LCA case contributing the arithmetic mean of its LM--LAD and LCX branch metrics.}
\label{tab:bspline-fitting-fidelity}
\scriptsize
\setlength{\tabcolsep}{4pt}
\begin{tabular*}{\textwidth}{@{\extracolsep{\fill}}ccccccccc@{}}
\toprule
$K_{\mathrm{cp}}$ &
\multicolumn{4}{c}{\textbf{Centreline fitting error (mm)} $\downarrow$} &
\multicolumn{4}{c}{\textbf{Vessel clDice (\%)} $\uparrow$} \\
\cmidrule(lr){2-5}\cmidrule(lr){6-9}
& \textbf{RCA} & \textbf{LM--LAD} & \textbf{LCX} & \textbf{Overall}
& \textbf{RCA} & \textbf{LM--LAD} & \textbf{LCX} & \textbf{Overall} \\
\midrule
8  & \metriccell{1.18}{0.72} & \metriccell{0.65}{0.40} & \metriccell{0.81}{0.38} & \metriccell{0.95}{0.60}
& \metriccell{86.21}{17.91} & \metriccell{96.98}{7.11} & \metriccell{92.78}{9.66} & \metriccell{90.55}{14.21} \\
12 & \metriccell{0.56}{0.41} & \metriccell{0.31}{0.21} & \metriccell{0.46}{0.26} & \metriccell{0.47}{0.33}
& \metriccell{97.35}{6.61} & \metriccell{99.47}{2.24} & \metriccell{98.01}{4.45} & \metriccell{98.04}{5.08} \\
20 & \metriccell{0.19}{0.13} & \metriccell{0.11}{0.08} & \metriccell{0.17}{0.11} & \metriccell{0.16}{0.11}
& \metriccell{99.88}{0.84} & \metriccell{100.00}{0.06} & \metriccell{99.94}{0.49} & \metriccell{99.92}{0.62} \\
\shortstack{Dense coordinates} & $0.00$ & $0.00$ & $0.00$ & $0.00$ & $100.00$ & $100.00$ & $100.00$ & $100.00$ \\
\bottomrule
\end{tabular*}
\end{table*}

\begin{table*}[t]
\centering
\caption{Effect of centreline representation on coarse and refined reconstruction using two input views. The dense-centreline variant directly predicts and refines the $N=200$ ordered centreline points, whereas the B-spline variant predicts and refines $K_{\mathrm{cp}}=20$ control points before decoding them to the same number of dense points. All other model and evaluation settings are unchanged. CD denotes Chamfer distance, and CC count denotes connected-component count. Values are mean $\pm$ standard error; \textbf{bold} indicates the better representation within each matched prediction stage.}
\label{tab:centreline-representation-ablation}
\scriptsize
\setlength{\tabcolsep}{2pt}
\begin{tabular*}{\textwidth}{@{\extracolsep{\fill}}
>{\centering\arraybackslash}m{0.08\textwidth}
>{\raggedright\arraybackslash}m{0.15\textwidth}
*{8}{>{\centering\arraybackslash}m{0.073\textwidth}}@{}}
\toprule
\multirow{2}{*}{\textbf{Stage}}
& \multirow{2}{*}{\shortstack[l]{\textbf{Centreline}\\\textbf{representation}}}
& \multicolumn{4}{c}{\textbf{RCA}}
& \multicolumn{4}{c}{\textbf{LCA}} \\
\cmidrule(lr){3-6}\cmidrule(lr){7-10}
&
& \shortstack{\textbf{Dice}\\\textbf{(\%)} $\uparrow$}
& \shortstack{\textbf{clDice}\\\textbf{(\%)} $\uparrow$}
& \shortstack{\textbf{CD}\\\textbf{(mm)} $\downarrow$}
& \shortstack{\textbf{CC}\\\textbf{count} $\downarrow$}
& \shortstack{\textbf{Dice}\\\textbf{(\%)} $\uparrow$}
& \shortstack{\textbf{clDice}\\\textbf{(\%)} $\uparrow$}
& \shortstack{\textbf{CD}\\\textbf{(mm)} $\downarrow$}
& \shortstack{\textbf{CC}\\\textbf{count} $\downarrow$} \\
\midrule
\multirow{2}{*}{Coarse} & Dense centreline
& \metriccell{10.22}{0.74} & \metriccell{10.23}{0.98} & \metriccell{13.81}{0.46} & \metriccell{\mathbf{1.284}}{0.056}
& \metriccell{8.97}{0.53} & \metriccell{8.29}{0.76} & \metriccell{16.09}{0.59} & \metriccell{2.276}{0.125} \\
& B-spline
& \metriccell{\mathbf{19.47}}{1.12} & \metriccell{\mathbf{20.40}}{1.56} & \metriccell{\mathbf{9.70}}{0.41} & \metriccell{1.324}{0.064}
& \metriccell{\mathbf{14.92}}{0.77} & \metriccell{\mathbf{13.94}}{0.90} & \metriccell{\mathbf{12.37}}{0.51} & \metriccell{\mathbf{1.605}}{0.097} \\
\midrule
\multirow{2}{*}{Refined} & Dense centreline
& \metriccell{22.18}{1.07} & \metriccell{26.87}{1.57} & \metriccell{10.36}{0.47} & \metriccell{1.189}{0.050}
& \metriccell{15.21}{0.68} & \metriccell{15.77}{0.91} & \metriccell{13.87}{0.58} & \metriccell{1.526}{0.087} \\
& B-spline
& \metriccell{\mathbf{66.92}}{1.50} & \metriccell{\mathbf{80.62}}{1.92} & \metriccell{\mathbf{3.16}}{0.27} & \metriccell{\mathbf{1.054}}{0.026}
& \metriccell{\mathbf{49.45}}{1.32} & \metriccell{\mathbf{61.30}}{1.93} & \metriccell{\mathbf{5.98}}{0.47} & \metriccell{\mathbf{1.053}}{0.026} \\
\bottomrule
\end{tabular*}
\end{table*}


\subsubsection{Coarse Model Encoder Details}
\label{suppsec:image-backbones}

This section provides the implementation details of the frozen VGGT encoder used by the coarse predictor in Subsection~\ref{subsec:coarse-predictor}.

\paragraph{VGGT backbone.}

The VGGT backbone~\citep{wang2025vggt} processes each input view independently and produces visual features
\begin{equation}
\bm{F}_k
=
f_{\mathrm{VGGT}}(\bm{I}_k),
\qquad
\bm{F}_k\in\mathbb{R}^{T_{\mathrm{VGGT}}\times d_{\mathrm{VGGT}}}.
\end{equation}
A learned projection and layer normalisation map these features to the common dimension:
\begin{equation}
\bm{H}^{I}_k
=
\operatorname{LN}
\left(
\bm{F}_k\bm{W}_{\mathrm{VGGT}}
+\bm{\beta}_{\mathrm{VGGT}}
\right)
\in
\mathbb{R}^{T_{\mathrm{VGGT}}\times d},
\end{equation}
where $\bm{W}_{\mathrm{VGGT}}\in\mathbb{R}^{d_{\mathrm{VGGT}}\times d}$ and $\bm{\beta}_{\mathrm{VGGT}}\in\mathbb{R}^{d}$ are learned. Under the shared notation of the main paper, $T_{\mathrm{img}}=T_{\mathrm{VGGT}}$.

To assess the effect of the image backbone, Table~\ref{tab:backbone-two-view-comparison} compares the frozen VGGT backbone~\citep{wang2025vggt} with ResNet-101~\citep{he2016resnet} using two input views. The ResNet alternative uses ResNet-101 without fine-tuning its backbone parameters. 

\begin{table*}[t]
\centering
\caption{Two-view backbone comparison for RCA and LCA coarse and refined reconstruction. CD denotes Chamfer distance, and CC count denotes connected-component count. Values are mean $\pm$ standard error.}
\label{tab:backbone-two-view-comparison}
\scriptsize
\setlength{\tabcolsep}{2pt}
\begin{tabular*}{\textwidth}{@{\extracolsep{\fill}}
>{\raggedright\arraybackslash}m{0.12\textwidth}
>{\raggedright\arraybackslash}m{0.09\textwidth}
*{8}{>{\centering\arraybackslash}m{0.075\textwidth}}@{}}
\toprule
\multirow{2}{*}{\textbf{Backbone}}
& \multirow{2}{*}{\textbf{Output}}
& \multicolumn{4}{c}{\textbf{RCA}}
& \multicolumn{4}{c}{\textbf{LCA}} \\
\cmidrule(lr){3-6}\cmidrule(lr){7-10}
&
& \shortstack{\textbf{Dice}\\\textbf{(\%)} $\uparrow$}
& \shortstack{\textbf{clDice}\\\textbf{(\%)} $\uparrow$}
& \shortstack{\textbf{CD}\\\textbf{(mm)} $\downarrow$}
& \shortstack{\textbf{CC}\\\textbf{count} $\downarrow$}
& \shortstack{\textbf{Dice}\\\textbf{(\%)} $\uparrow$}
& \shortstack{\textbf{clDice}\\\textbf{(\%)} $\uparrow$}
& \shortstack{\textbf{CD}\\\textbf{(mm)} $\downarrow$}
& \shortstack{\textbf{CC}\\\textbf{count} $\downarrow$} \\
\midrule
ResNet-101 & Coarse
& \metriccell{18.94}{1.15} & \metriccell{19.94}{1.43} & \metriccell{9.37}{0.37} & \metriccell{1.428}{0.097}
& \metriccell{13.88}{1.78} & \metriccell{13.69}{0.89} & \metriccell{12.99}{0.32} & \metriccell{1.878}{0.097} \\
\cmidrule(lr){2-10}
ResNet-101 & Refined
& \metriccell{60.62}{1.29} & \metriccell{72.80}{1.61} & \metriccell{4.06}{0.47} & \metriccell{1.087}{0.029}
& \metriccell{45.19}{1.71} & \metriccell{56.33}{1.50} & \metriccell{6.71}{0.44} & \metriccell{1.102}{0.064} \\
\midrule
VGGT & Coarse
& \metriccell{19.47}{1.12} & \metriccell{20.40}{1.56} & \metriccell{9.70}{0.41} & \metriccell{1.324}{0.064}
& \metriccell{14.92}{0.77} & \metriccell{13.94}{0.90} & \metriccell{12.37}{0.51} & \metriccell{1.605}{0.097} \\
\cmidrule(lr){2-10}
VGGT & Refined
& \metriccell{66.92}{1.50} & \metriccell{80.62}{1.92} & \metriccell{3.16}{0.27} & \metriccell{1.054}{0.026}
& \metriccell{49.45}{1.32} & \metriccell{61.30}{1.93} & \metriccell{5.98}{0.47} & \metriccell{1.053}{0.026} \\
\bottomrule
\end{tabular*}
\end{table*}

\subsubsection{Branch Decoder Details}
\label{suppsec:branch-decoder}

The parallel coarse decoder uses $M$ trainable branch queries $\bm{q}_m\in\mathbb{R}^{d}$ and the query matrix $\bm{Q}_{\mathrm{query}}=[\bm{q}_1;\ldots;\bm{q}_M]$. Every query attends to the multiview memory but not to another branch query. Using $\bm{E}_{\mathrm{view}}$ as the cross-attention keys and values, the shared decoder produces branch tokens
\begin{equation}
[\bm{b}_1;\ldots;\bm{b}_M]
=
f_{\mathrm{cross\text{-}attn}}
(\bm{Q}_{\mathrm{query}},\bm{E}_{\mathrm{view}}),
\qquad
\bm{b}_m\in\mathbb{R}^{d}.
\end{equation}
The branch queries are decoded independently and in parallel and correspond to the ordered target entries defined in Subsection~\ref{subsec:dataset}. Query $m$ therefore retrieves evidence for branch entry $m$, but optional entries do not denote fixed, named anatomical side-branch categories. An existence prediction determines which optional branch outputs are active for an individual reconstruction. The prediction of side branch $m$ is directly supervised by target entry $m$; no Hungarian matching is used.

The control head is
\begin{equation}
g_{\mathrm{cp}}:
\mathbb{R}^{d}\rightarrow
\mathbb{R}^{3K_{\mathrm{cp}}},
\qquad
\hat{\bm{P}}^{(0)}_m
=
\operatorname{reshape}(g_{\mathrm{cp}}(\bm{b}_m)).
\end{equation}
The raw radius head predicts $N$ log-radii,
\begin{equation}
g_{\mathrm{rad}}:
\mathbb{R}^{d}\rightarrow\mathbb{R}^{N},
\qquad
\hat{\bm{r}}^{(0)}_m
=
\exp(g_{\mathrm{rad}}(\bm{b}_m)),
\end{equation}
with the log values clamped to a finite numerical range before exponentiation. The branch-existence head produces one logit per query,
\begin{equation}
g_{\mathrm{exist}}:
\mathbb{R}^{d}\rightarrow\mathbb{R},
\qquad
z_{m}=g_{\mathrm{exist}}(\bm{b}_{m}).
\end{equation}
At inference, the main-branch entry is forced active and side-branch probabilities are $\hat e_{m}=\sigma(z_{m})$. The prediction heads are shared across all branch queries, and every output is decoded in the same world frame. Thus, $M$ is a fixed maximum number of query outputs, while the predicted active branch count can vary between cases. We use maximum query counts of $M_{\mathrm{RCA}}=7$ and $M_{\mathrm{LCA}}=13$. The required primary branches are always active. An optional branch is retained at inference when $\sigma(z_{m}) \geq 0.5$.

\subsubsection{Projection-Guided Geometry and Radius Refiners}
\label{suppsec:geometry-refiner}
\label{suppsec:refiners}

\paragraph{B-spline geometry refinement.}

The coarse predictor supplies a population-level artery estimate, but ambiguity in the sparse input views can leave case-specific centreline discrepancies. The geometry refiner corrects these discrepancies directly in the compact B-spline control-point space while holding the coarse radii $\hat{\bm{r}}^{(0)}_m$ fixed. Its inputs are the coarse controls $\hat{\bm{P}}^{(0)}$, branch tokens, coronary mask images, acquisition directions, spatial image features extracted by the frozen backbone, and the known differentiable projection $\Pi_k:\mathbb{R}^{3}\rightarrow\mathbb{R}^{2}$ for each view $k$. The same frozen-backbone features used by the coarse predictor are therefore reused during geometry refinement; the backbone itself remains frozen.

Because a B-spline control point does not generally lie on the decoded curve, each control index $j$ is associated with an on-curve anchor. The anchor index is the sampled centreline position at which the corresponding basis function is strongest:
\begin{equation}
n_j=\arg\max_{n}A_{n,j}.
\end{equation}
At refinement stage $s$, the current decoded anchor is
\begin{equation}
\bm{a}^{(s)}_{m,j}
=
\sum_{\ell=1}^{K_{\mathrm{cp}}}
A_{n_j,\ell}\hat{\bm{p}}^{(s)}_{m,\ell}.
\end{equation}
Here, $s\in\{0,\ldots,S_g-1\}$ denotes the geometry-refinement stage. The anchor is projected to view $k$ as $\bm{u}^{(s)}_{m,j,k}=\Pi_k(\bm{a}^{(s)}_{m,j})$. Around $\bm{u}^{(s)}_{m,j,k}$, the implementation samples an odd-sized input-mask patch, local learned image features, the mask distance transform, and its two-dimensional gradient. It also extracts the backbone feature at the corresponding image region: from the spatial VGGT patch-token map in the final model, or from the ResNet feature map in the backbone comparison. Collectively, these samples form the local multiview projection evidence for control point $(m,j)$. Samples outside the detector or belonging to padded views are marked invalid.

The evidence encoder maps the local samples to dimension $d$. A refinement query combines the current control position, anchor position, branch token, control-index embedding, branch-index embedding, and recurrent state. Multi-head attention aggregates the encoded samples over valid views. The resulting state is passed to a residual head:
\begin{equation}
\Delta\hat{\bm{p}}^{(s)}_{m,j}
=
\alpha_g
\tanh
\left(
g_g(\bm{z}^{(s)}_{m,j})
\right)
\in\mathbb{R}^{3},
\end{equation}
where $\bm{z}^{(s)}_{m,j}\in\mathbb{R}^{d}$ is the evidence-conditioned state, $g_g:\mathbb{R}^{d}\rightarrow\mathbb{R}^{3}$ is the learned residual head, and $\alpha_g$ is the maximum update magnitude in millimetres. The controls are updated by
\begin{equation}
\hat{\bm{P}}^{(s+1)}
=
\hat{\bm{P}}^{(s)}
+
\Delta\hat{\bm{P}}^{(s)}.
\end{equation}
The output layer parameters are shared across the configured $S_g$ stages. After each stage, the updated B-spline is decoded, its anchors are reprojected, and new image evidence is sampled for the next stage. The radius vector remains fixed throughout this module. Geometry refinement is therefore a fixed-depth learned computation at inference rather than per-case gradient-based optimisation.

The reported geometry-refiner configuration uses $S_g=4$ stages, $17\times17$ local patches, a 256-dimensional evidence state, and a maximum per-stage control update of $\alpha_g=7\,\mathrm{mm}$.


\paragraph{Dense-radius refinement.}
\label{suppsec:radius-refiner}

The radius refiner is enabled only with B-spline centrelines and raw dense-radius prediction. It receives the geometry-refined vessel, holds the refined centreline $\hat{\bm{C}}^{(S_g)}_m$ fixed, and updates only the dense radii. At radius-refinement stage $u\in\{0,\ldots,S_r-1\}$, the current centreline-and-radius model is differentiably rendered into every valid input view.

For projected centreline point $(m,n)$, finite differences along the projected curve estimate a unit tangent. Rotating this tangent by $90^{\circ}$ gives the transverse image-plane normal. Let $L$ be the odd number of profile samples and $w_{\mathrm{prof}}$ the profile half-width in pixels. The sampling grid is
\begin{equation}
\bm{u}_{m,n,k,\ell}
=
\Pi_k(\hat{\bm{c}}_{m,n})
+
\delta_{\ell}\bm{n}_{m,n,k},
\qquad
\delta_{\ell}\in[-w_{\mathrm{prof}},w_{\mathrm{prof}}],
\quad
\ell\in\{1,\ldots,L\},
\end{equation}
where $\bm{n}_{m,n,k}$ is the transverse unit normal. Four profiles are sampled: the observed input mask, the rendered vessel mask, their signed difference, and their absolute difference. These profiles expose whether the rendered vessel is too narrow or too wide around the current projected centreline while reducing interference from distant parts of the projected coronary tree. They are concatenated with the encoded view information and mapped to the model dimension.

For point $(m,n)$, the query is formed from the branch token, branch-index and centreline-point-index embeddings, encoded current radius, and recurrent point state. Multi-head attention aggregates the valid per-view profiles. Two one-dimensional convolutional blocks then propagate context along the ordered centreline. A zero-initialised scalar residual head predicts
\begin{equation}
\Delta\hat r^{(u)}_{m,n}
=
\alpha_r
\tanh
\left(
g_r(\bm{z}^{(u)}_{m,n})
\right),
\end{equation}
where $u$ is the radius-refinement stage, $\bm{z}^{(u)}_{m,n}\in\mathbb{R}^{d}$ is the evidence-conditioned point state, $g_r:\mathbb{R}^{d}\rightarrow\mathbb{R}$ is the residual head, and $\alpha_r$ bounds one correction in millimetres. The update is
\begin{equation}
\hat r^{(u+1)}_{m,n}
=
\max
\left(
r_{\min},
\hat r^{(u)}_{m,n}+\Delta\hat r^{(u)}_{m,n}
\right).
\end{equation}
The updated surface is rerendered and new transverse evidence is sampled before the next of the $S_r$ stages. Because centreline coordinates are detached from this pathway, radius refinement cannot change vessel trajectory to reduce its loss. After the configured geometry and radius stages, the final refined vessel code is
\begin{equation}
\hat{\bm{V}}_{\mathrm{refined}}
=
\left\{
(\hat{\bm{c}}^{(S_g)}_{m,n},
\hat r^{(S_r)}_{m,n})
\right\}_{m,n}.
\end{equation}

The reported radius-refiner configuration uses $S_r=3$ stages, $L=25$ transverse samples spanning $w_{\mathrm{prof}}=16$ pixels on either side of the projected centreline, a maximum update of $\alpha_r=1\,\mathrm{mm}$ per stage, and minimum radius $r_{\min}=0.05\,\mathrm{mm}$.

\subsubsection{Within-Case Branch-Subset Training}
\label{suppsec:training-signals}

Branch-count variation in the original dataset is entangled with differences in case-specific anatomy. To provide direct supervision for optional branch presence, we construct multiple branch-subset variants from each training case while holding its underlying anatomy fixed.

For a case with $B$ ordered branches, every variant retains the required main path or paths: branch 1 for RCA and branches 1--2 for LCA. Starting from the complete artery tree, optional side branches are successively removed in reverse canonical order until only the required main path or paths remain. An RCA case therefore produces $B$ variants, while an LCA case produces $B-1$ variants, including the complete-tree and main-path-only versions. For every variant, we regenerate the projected masks and branch-existence labels while preserving the geometry targets and ordered identities of retained branches.

All variants from the same case are collated in one optimisation batch and use the same selected input views. The training loader visits every physical case once per epoch, and every corresponding branch-subset variant is included; no Bernoulli sampling probability is used. The variants are presented together rather than introduced sequentially as a curriculum. This protocol is applied when training both the coarse predictor and the geometry refiner.

\begin{table*}[t]
\centering
\caption{Effect of branch-subset augmentation on two-view coarse prediction and geometry refinement. CD denotes Chamfer distance, and CC count denotes connected-component count. Values are mean $\pm$ standard error.}
\label{tab:progressive-branch-augmentation}
\scriptsize
\setlength{\tabcolsep}{2pt}
\begin{tabular*}{\textwidth}{@{\extracolsep{\fill}}
>{\raggedright\arraybackslash}m{0.14\textwidth}
>{\centering\arraybackslash}m{0.09\textwidth}
*{8}{>{\centering\arraybackslash}m{0.072\textwidth}}@{}}
\toprule
\multirow{2}{*}{\textbf{Prediction stage}}
& \multirow{2}{*}{\shortstack{\textbf{Branch-subset}\\\textbf{augmentation}}}
& \multicolumn{4}{c}{\textbf{RCA}}
& \multicolumn{4}{c}{\textbf{LCA}} \\
\cmidrule(lr){3-6}\cmidrule(lr){7-10}
&
& \shortstack{\textbf{Dice}\\\textbf{(\%)} $\uparrow$}
& \shortstack{\textbf{clDice}\\\textbf{(\%)} $\uparrow$}
& \shortstack{\textbf{CD}\\\textbf{(mm)} $\downarrow$}
& \shortstack{\textbf{CC}\\\textbf{count} $\downarrow$}
& \shortstack{\textbf{Dice}\\\textbf{(\%)} $\uparrow$}
& \shortstack{\textbf{clDice}\\\textbf{(\%)} $\uparrow$}
& \shortstack{\textbf{CD}\\\textbf{(mm)} $\downarrow$}
& \shortstack{\textbf{CC}\\\textbf{count} $\downarrow$} \\
\midrule
Coarse prediction & \xmark
& \metriccell{11.38}{0.95} & \metriccell{10.89}{1.09} & \metriccell{13.42}{0.52} & \metriccell{1.338}{0.070}
& \metriccell{9.82}{0.54} & \metriccell{8.81}{0.59} & \metriccell{14.70}{0.49} & \metriccell{1.882}{0.082} \\
Coarse prediction & \cmark
& \metriccell{19.47}{1.12} & \metriccell{20.40}{1.56} & \metriccell{9.70}{0.41} & \metriccell{1.324}{0.064}
& \metriccell{14.92}{0.77} & \metriccell{13.94}{0.90} & \metriccell{12.37}{0.51} & \metriccell{1.605}{0.097} \\
\midrule
Geometry refiner & \xmark
& \metriccell{46.57}{1.45} & \metriccell{58.45}{2.04} & \metriccell{6.13}{0.43} & \metriccell{1.068}{0.029}
& \metriccell{37.52}{1.04} & \metriccell{44.57}{1.51} & \metriccell{8.01}{0.45} & \metriccell{1.211}{0.054} \\
Geometry refiner & \cmark
& \metriccell{65.89}{1.44} & \metriccell{80.84}{1.90} & \metriccell{3.16}{0.27} & \metriccell{1.068}{0.029}
& \metriccell{48.43}{1.29} & \metriccell{61.38}{1.93} & \metriccell{5.98}{0.47} & \metriccell{1.053}{0.026} \\
\bottomrule
\end{tabular*}
\end{table*}

Table~\ref{tab:progressive-branch-augmentation} shows that within-case branch-subset training improves Dice, clDice, and Chamfer distance for both prediction stages and artery cohorts. For branch-existence prediction, it increases accuracy from $92.95\%$ to $94.63\%$ and F1 from $88.37\%$ to $92.48\%$. The improvement is driven primarily by higher precision and specificity, indicating fewer false-positive side-branch predictions, although recall decreases modestly. These results suggest that varying branch presence while preserving patient anatomy helps the model learn that side branches are optional.

\subsubsection{Direct Per-Scene B-spline Optimisation Baseline}
\label{suppsec:direct-optimization}

The direct-optimisation baseline starts from the same coarse control points $\hat{\bm{P}}^{(0)}$ as the learned geometry refiner. For each test case, gradient-based optimisation updates only the B-spline controls to reduce the selected 2D projection objective against the input masks:
\begin{equation}
\bm{P}^{\star}_{\mathrm{opt}}
=
\arg\min_{\bm{P}}
\mathcal{L}_{\mathrm{proj}}
\left(
\operatorname{Render}(\bm{A}\bm{P},\hat{\bm{r}}^{(0)}),
\{\bm{I}_k\}_{k=1}^{K}
\right).
\end{equation}
The coarse radius is fixed. The number of optimisation iterations and all hyperparameters are determined during method development and fixed before test-set evaluation. This baseline isolates whether the learned geometry refiner provides a better accuracy--runtime trade-off than test-time optimisation from the same initialisation.

Table~\ref{tab:refinement-ablation} compares direct optimisation with the learned geometry and radius refiners from a common two-view coarse initialisation for RCA and LCA, including their post-coarse processing time.

\subsection{Implementation Details and Ablation Studies}
\label{suppsec:implementation-and-ablations}

\subsubsection{Experiment Details}
\label{suppsec:experiment-details}

PPCAR-Net is implemented in PyTorch and trained with one GPU per run. We use AdamW with an initial learning rate of $10^{-4}$, weight decay of $10^{-4}$, cosine learning-rate decay to $10^{-6}$, and a gradient-norm clipping threshold of $1.0$. The coarse predictor, geometry refiner, and radius refiner all use a batch size of 16. The coarse predictor and geometry refiner are trained for at most 200 epochs, and the radius refiner for at most 100 epochs. The frozen image backbone is initialised from the pretrained \texttt{facebook/VGGT-1B} checkpoint, and its features are precomputed before predictor training. The test set is evaluated only after model and hyperparameter selection. All reported runtime measurements use a single NVIDIA GeForce RTX 4090 GPU.

\subsubsection{Evaluation Implementation Details}
\label{suppsec:evaluation-implementation}

All volumetric metrics are computed on the common isotropic $0.5\,\mathrm{mm}$ evaluation grid defined in Subsection~\ref{subsec:evaluation-metrics}. The ground-truth mask is resampled by nearest-neighbour interpolation, while the predicted centreline and radii are rasterised as the union of filled, radius-varying tubular segments on this grid. The resulting voxelised prediction is used consistently for Dice, clDice, and connected-component counting.

Let $\widehat{\mathcal{V}}$ and $\mathcal{V}$ denote the foreground-voxel sets of the predicted and ground-truth arteries, respectively. Their volumetric overlap is measured by
\begin{equation}
\operatorname{Dice}
=
\frac{2\lvert\widehat{\mathcal{V}}\cap\mathcal{V}\rvert}
{\lvert\widehat{\mathcal{V}}\rvert+\lvert\mathcal{V}\rvert}.
\end{equation}
For clDice, let $\mathcal{S}(\cdot)$ denote the one-voxel 3D centreline obtained by skeletonisation. Topology precision, topology sensitivity, and clDice are
\begin{equation}
\begin{aligned}
T_{\mathrm{prec}}
&=
\frac{\lvert\mathcal{S}(\widehat{\mathcal{V}})\cap\mathcal{V}\rvert}
{\lvert\mathcal{S}(\widehat{\mathcal{V}})\rvert},
\\
T_{\mathrm{sens}}
&=
\frac{\lvert\mathcal{S}(\mathcal{V})\cap\widehat{\mathcal{V}}\rvert}
{\lvert\mathcal{S}(\mathcal{V})\rvert},
\\
\operatorname{clDice}
&=
\frac{2T_{\mathrm{prec}}T_{\mathrm{sens}}}
{T_{\mathrm{prec}}+T_{\mathrm{sens}}}.
\end{aligned}
\end{equation}
Each ground-truth RCA or LCA tree contains exactly one connected component because all retained side branches attach to the corresponding main branch. Connected-component count evaluates whether a prediction preserves this expected structure. It is computed from the same $0.5\,\mathrm{mm}$ voxelised prediction used for Dice and clDice. We apply 3D connected-component labelling to the foreground set $\widehat{\mathcal{V}}$ and report
\begin{equation}
N_{\mathrm{CC}}
=
\left\lvert\operatorname{CC}(\widehat{\mathcal{V}})\right\rvert,
\end{equation}
where $\operatorname{CC}(\widehat{\mathcal{V}})$ is the set of connected foreground components. A value of one indicates a single connected prediction, while larger values indicate fragmentation.

Finally, let $\widehat{\mathcal{C}}$ and $\mathcal{C}$ be the predicted and ground-truth centreline point sets in millimetre-valued world coordinates. We use the symmetric Chamfer distance
\begin{equation}
\operatorname{CD}(\widehat{\mathcal{C}},\mathcal{C})
=
\frac{1}{\lvert\widehat{\mathcal{C}}\rvert}
\sum_{\hat{\bm c}\in\widehat{\mathcal{C}}}
\min_{\bm c\in\mathcal{C}}\lVert\hat{\bm c}-\bm c\rVert_2
+
\frac{1}{\lvert\mathcal{C}\rvert}
\sum_{\bm c\in\mathcal{C}}
\min_{\hat{\bm c}\in\widehat{\mathcal{C}}}\lVert\bm c-\hat{\bm c}\rVert_2.
\end{equation}
Dice and clDice are reported as percentages, Chamfer distance is reported in millimetres, and connected-component count is reported as a unitless count.

\subsubsection{Comparison-Method Adaptation, Evaluation, and Timing}
\label{suppsec:comparison-method-implementation}

We adapted the available implementations of 3DGR-CAR~\citep{fu20243dgrcar}, AutoCAR~\citep{zhu2025autocar}, and DeepCA~\citep{wang2025deepca} to the same ImageCAS-derived data protocol used by our method. This experiment is not intended to reproduce the results reported on the original datasets of those papers. Instead, we train their models on our dataset and evaluate them under the same data split, input-view protocol, and metrics as our method to provide a controlled comparison. The available 3DGR-CAR and AutoCAR code did not contain every module required to train and evaluate their complete pipelines on our dataset. We therefore reimplemented the missing components from the algorithmic and implementation descriptions in the corresponding papers. The remaining adaptations principally concern the data-loading and evaluation interfaces. Each comparison-method loader reads the segmented two-dimensional coronary masks, their associated acquisition geometry, and the corresponding three-dimensional artery label, then converts these inputs to the tensor layout and spatial resolution expected by that method. The same training, validation, and test case assignments are used across methods. For each supported input-view count, all methods use the same selected views for a given test case so that differences are not caused by a different choice of projections.

Each comparison model is trained from scratch on our training split so that the experiment evaluates model capacity under the same data and supervision, without an advantage from external pre-training. Learning rate, training duration, loss weights, output threshold, and other method-specific settings are selected using reconstruction performance on the validation split. The test split is not used for model selection or hyperparameter tuning. Apart from modules needed to load our data or complete unavailable output and evaluation steps from the descriptions in the corresponding papers, the published model structure and learning objective are retained.

All final predictions are evaluated by the same code. Volumetric outputs are resampled to the common isotropic $0.5$ mm grid and binarised using the validation-selected threshold before Dice and clDice are calculated. The native outputs have different spatial dimensions and voxel resolutions: AutoCAR produces a $400^3$ volume at $0.5$ mm resolution, whereas DeepCA and 3DGR-CAR produce $128^3$ volumes at approximately $0.9167$ mm and $0.9239$ mm resolution, respectively. Consequently, resampling each prediction to the common grid can introduce method-dependent discretisation effects, so the Dice and clDice comparison is not entirely independent of native output resolution. Nevertheless, these dimensions are part of the original output configuration of each model; we retain them rather than altering the model architectures or retraining them for a common output grid. When a method does not natively produce an ordered centreline, its predicted volume is converted to a vascular graph using the same graph-extraction post-processing, and Chamfer distance is then computed between its extracted centreline point set and the ground-truth centreline point set in millimetres. Reported values are the mean and standard error across the test cases. This shared evaluator keeps the metric definitions, physical coordinate convention, and test cases fixed across all methods.

\paragraph{Inference-time breakdown.}
\label{suppsec:inference-time}

All runtimes in Table~\ref{tab:supp-inference-time-results} use two input views and a single NVIDIA GeForce RTX 4090 GPU under the same software setup. We report \emph{model inference time}, measured from supplying the 2D input views to obtaining the method's native reconstruction output; \emph{post-processing time}, measured from the native output to a vascular graph/skeleton representation using the graph-extraction algorithm of AutoCAR~\citep{zhu2025autocar}; and their sum as the \emph{total time}. This post-processing is performed for every quantitatively evaluated case so that the final metrics use a graph/skeleton representation. Although the same graph-extraction algorithm is applied to all comparison methods, its runtime depends on the native input representation, evaluation-grid size, and vessel-mask complexity. DeepCA supplies a compact $128^3$ binary \texttt{uint8} mask without probability thresholding and therefore has the shortest post-processing time. The $128^3$ binary output of 3DGR-CAR is accompanied by additional arrays, requiring more input/output validation and potentially presenting a more complex foreground. AutoCAR usually supplies a substantially larger $400^3$ probability volume, which incurs additional decompression, validation, and thresholding and consequently has the longest post-processing time.

\begin{table*}[t]
\centering
\caption{Two-view reconstruction-time comparison on one NVIDIA GeForce RTX 4090 GPU. Model inference produces each method's native output, post-processing converts that output into a vascular graph/skeleton representation when required, and total time is their sum. The best result is shown in \textbf{bold} and the second-best result is \underline{underlined}; post-processing times are ranked only among methods that require this stage. -- denotes that post-processing is not required; N/R denotes that it was not recorded.}
\label{tab:supp-inference-time-results}
\small
\setlength{\tabcolsep}{6pt}
\begin{tabular}{lccc}
\toprule
\textbf{Method} & \shortstack{\textbf{Model inference}\\\textbf{time} $\downarrow$} & \shortstack{\textbf{Post-processing}\\\textbf{time} $\downarrow$} & \shortstack{\textbf{Total}\\\textbf{time} $\downarrow$} \\
\midrule
SDF-CAR~\citep{reda2026sdfcar} & 45 min & N/R & N/R \\
3DGR-CAR~\citep{fu20243dgrcar} & 61 s & 2.71 s & 63.7 s \\
DeepCA~\citep{wang2025deepca} & 700 ms & 1.99 s & 2.69 s \\
AutoCAR~\citep{zhu2025autocar} & 330 ms & 3.43 s & 3.76 s \\
PPCAR-Net (coarse) & \textbf{60 ms} & -- & \textbf{0.06 s} \\
PPCAR-Net (refined) & \underline{121 ms} & -- & \underline{0.12 s} \\
\bottomrule
\end{tabular}
\end{table*}

\subsubsection{Ablation Studies}
\label{suppsec:ablation-studies}
With two input views, the branch-existence head achieves an accuracy of $94.63\%$ and an F1 score of $92.48\%$, indicating reliable identification of the branches present in each case. Table~\ref{tab:centreline-representation-ablation} compares direct dense-centreline prediction with the B-spline parameterisation. The B-spline representation improves Dice, clDice, and Chamfer distance at both prediction stages for RCA and LCA. It also lowers the refined connected-component count for both arteries and the coarse count for LCA; the dense-centreline variant has a marginally lower RCA coarse count. Table~\ref{tab:backbone-two-view-comparison} compares the frozen image encoders. After refinement, VGGT improves evaluation metrics over ResNet-101 for both RCA and LCA. At the coarse stage, VGGT improves Dice, clDice, and connected-component count for both arteries and Chamfer distance for LCA, while RCA Chamfer distance is slightly lower with ResNet-101. Table~\ref{tab:progressive-branch-augmentation} evaluates within-case branch-subset augmentation. This augmentation improves evaluation metrics for both prediction stages and artery cohorts. These results support using the variants to learn optional side-branch structure. Table~\ref{tab:supp-inference-time-results} separates native model inference from graph-extraction post-processing. Our native centreline-and-radius output requires no graph-conversion stage and completes refined two-view reconstruction in $0.12$ s. The refinement comparison is presented in Section~\ref{subsec:refinement-results} and Table~\ref{tab:refinement-ablation}.

\end{document}